%% file: main.tex
\documentclass[10pt,twocolumn,letterpaper]{article}

\usepackage[pagenumbers]{cvpr} 

\usepackage{algorithm}
\usepackage{algorithmic}

\usepackage{amsthm} 
\theoremstyle{plain}      
\newtheorem{theorem}{Theorem}[section]  
\newtheorem{lemma}[theorem]{Lemma}      %
\newtheorem{proposition}{Proposition}
\usepackage{adjustbox}

\definecolor{cvprblue}{rgb}{0.21,0.49,0.74}
\usepackage[pagebackref,breaklinks,colorlinks,allcolors=cvprblue]{hyperref}

\def\paperID{18639} 
\def\confName{CVPR}
\def\confYear{2026}

\title{EIRES:Training-free AI-Generated Image Detection via Edit-Induced Reconstruction Error Shift}

\author{
Wan Jiang$^{1}$ \qquad
Jing Yan$^{2}$ \qquad
Xiaojing Chen$^{2}$ \qquad
Ling Shen$^{2}$\\
Chenhao Lin$^{3}$ \qquad
Yunfeng Diao$^{1}$ \qquad
Richang Hong$^{1}$ \\
$^{1}$HFUT\qquad
 $^{2}$AHU \qquad $^{3}$XJTU
}

\begin{document}
\maketitle
\input{sec/0_Abstract}    
\input{sec/1_Intro}
\input{sec/2_Related}
\input{sec/3_Method}
\input{sec/4_Theory}
\input{sec/5_Experiment}
\input{sec/6_Conclusion}
\input{sec/X_suppl}
{   \clearpage
    \small
    \bibliographystyle{ieeenat_fullname}
    \bibliography{main}
}


\end{document}

%% file: sec/0_abstract.tex
\begin{abstract}
Diffusion models have recently achieved remarkable photorealism, making it increasingly difficult to distinguish real images from generated ones, raising significant privacy and security concerns. In response, we present a key finding: structural edits enhance the reconstruction of real images while degrading that of generated images, creating a distinctive edit-induced reconstruction error shift. This asymmetric shift enhances the separability between real and generated images.
Building on this insight, we propose EIRES, a training-free method that leverages structural edits to reveal inherent differences between real and generated images. To explain the discriminative power of this shift, we derive the reconstruction error lower bound under edit perturbations. Since EIRES requires no training, thresholding depends solely on the natural separability of the signal, where a larger margin yields more reliable detection. Extensive experiments show that EIRES is effective across diverse generative models and remains robust on the unbiased subset, even under post-processing operations.

\end{abstract}

%% file: sec/1_intro.tex
\section{Introduction}
\label{sec:intro}
The rapid development of large-scale generative models has greatly lowered the barrier to producing high-quality, photorealistic images. While this technological advancement fosters unprecedented creativity and innovation, it also introduces profound risks of misinformation, visual deception, and the erosion of public trust.
These risks have already materialized in multiple high-profile incidents, such as AI-generated images of President Biden being hospitalized or explicit deepfakes of singer Taylor Swift circulating widely on social media~\cite{reuters2024_fakepolitics, cbs2024_taylor_deepfake}, attracting tens of millions of views and sparking political and privacy concerns .
Such incidents underscore the urgent need for reliable, scalable, and generalizable methods to detect and trace AI-generated imagery, which are essential for preserving digital authenticity and sustaining societal trust.

\begin{figure}[t!]
\begin{center}
\includegraphics[width=1\linewidth]{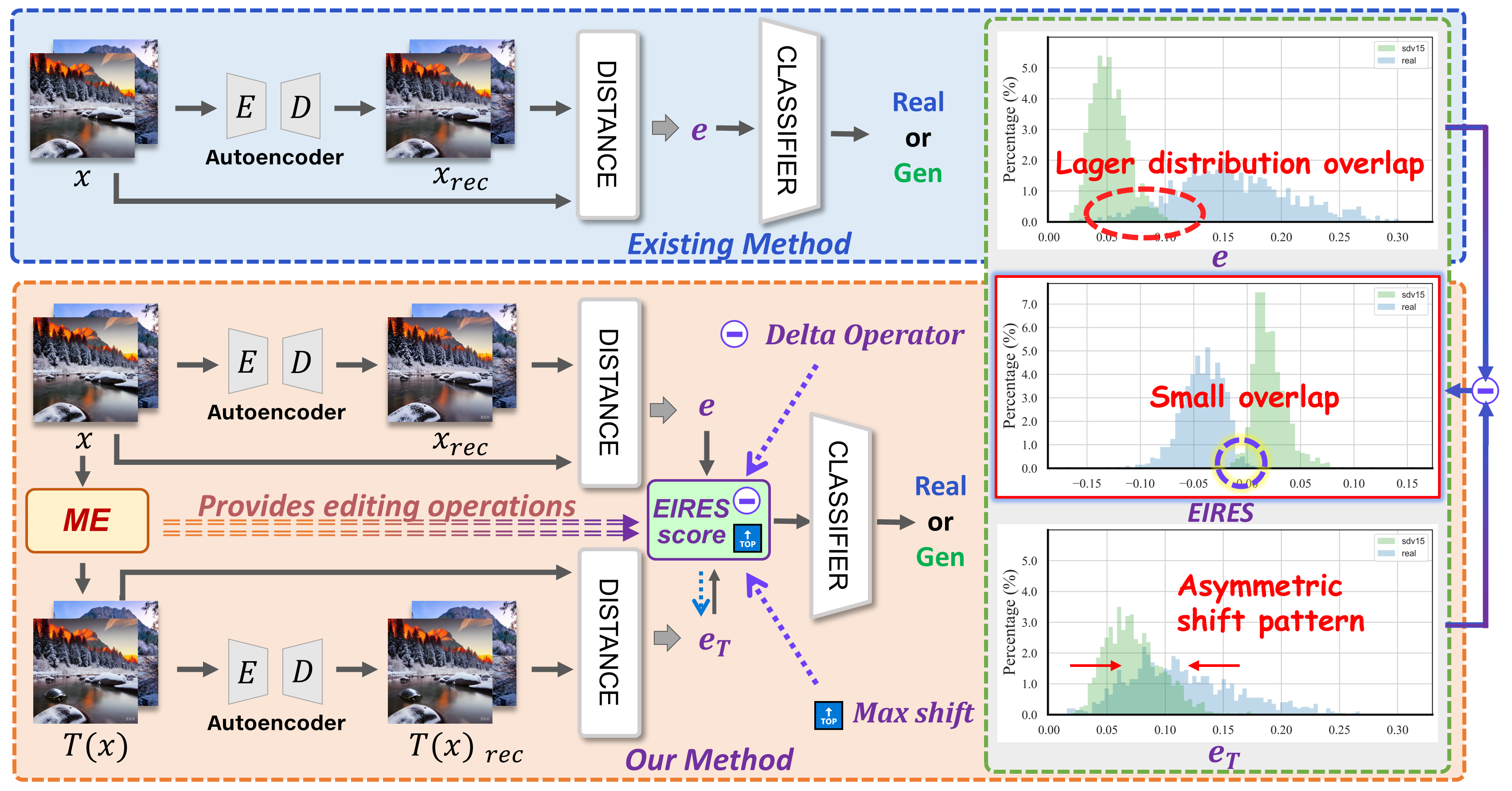}
\end{center}
\vspace{-4mm}
\caption{Comparison between reconstruction-based detector and EIRES.
Existing methods \cite{Aero} operate on raw reconstruction error, yielding a small real–fake margin and unstable thresholding. Our method applies multi-structured edits (ME) and computes the edit-induced max reconstruction error shift score via EIRES, producing a substantially larger margin for more reliable detection.}
\label{fig:vs}
\vspace{-4mm}
\end{figure}
\begin{figure*}[t!]
\begin{center}
\includegraphics[width=0.8\linewidth]{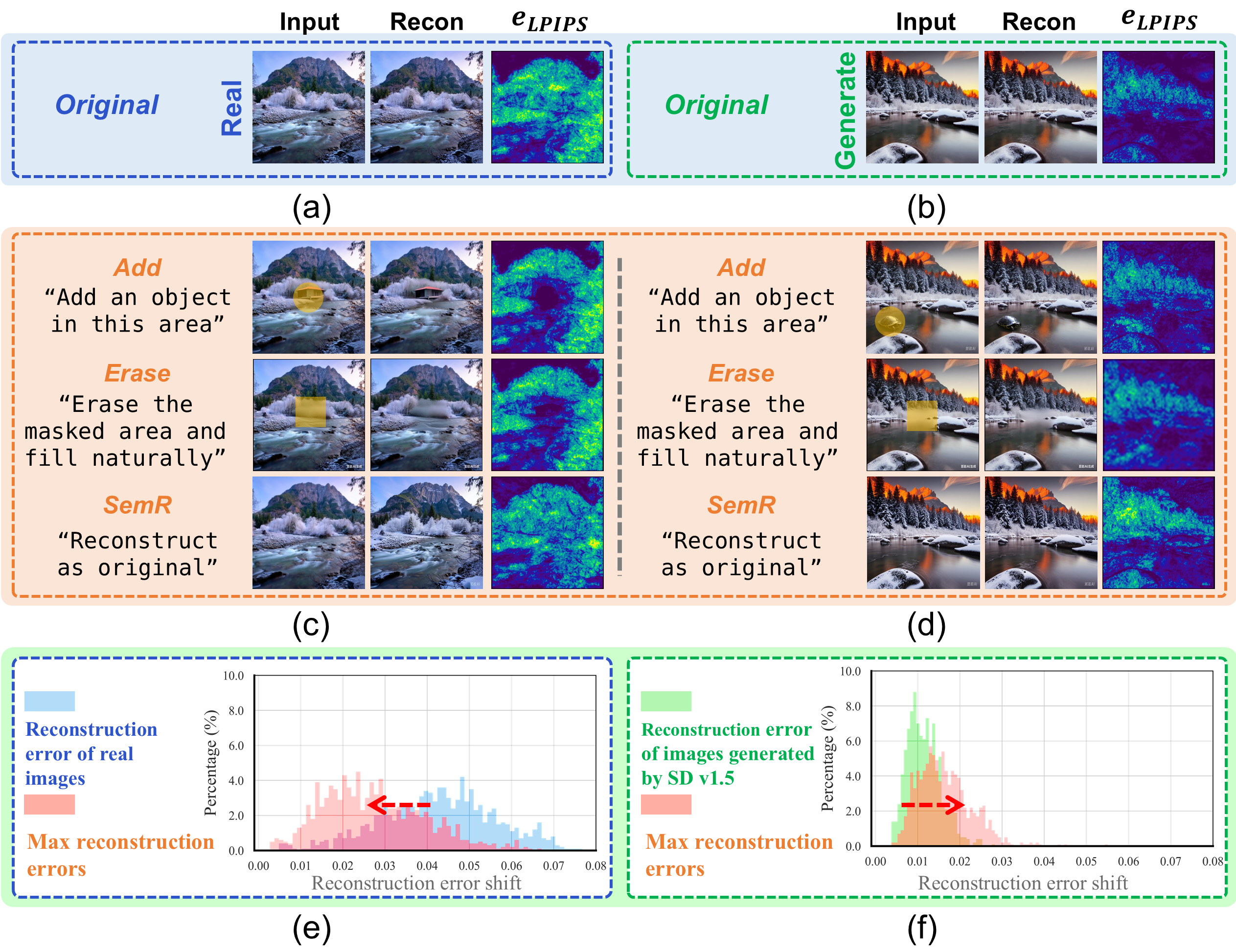}
\end{center}
\vspace{-4mm}
\caption{\textbf{Analysis of various image editing results and reconstruction errors shift.} (a) and (b) show real images from ImageNet~\cite{imagenet}  and their generated counterparts produced by a pre-trained Stable Diffusion~\cite{ CompVis2023StableDiffusion}, along with their autoencoder reconstructions and the corresponding LPIPS~\cite{zhang2018unreasonable} reconstruction error heatmaps. (c) and (d) display the results of applying different editing operations: Add, Erase and SemR to real and generated images, with the respective LPIPS reconstruction error heatmaps. (e) and (f) show the distribution of maximum reconstruction errors before and after applying the editing operations on real images from ImageNet and generated images from Stable Diffusion v1.5~\citep{rombach2022high}. As observed in (e), the reconstruction error for real images decreases after editing, while in (f) the error increases for generated images after editing. Notably, as shown in prior work~\cite{Aero}, the reconstruction error for generated images is generally lower than for real images.
}
\label{fig:moti}
\vspace{-1em}
\end{figure*}

Despite recent progress in detection methods, building reliable and generalizable detectors for AI-generated images remains a highly challenging task.
Reconstruction-based approaches \cite{Dire, SeDID, Aero, HFI} have recently gained significant attention for their flexibility and model-agnostic nature. These methods exploit raw reconstruction errors derived from autoencoders or diffusion models to distinguish real and generated images, under the assumption that real images are inherently more difficult to reconstruct faithfully~\cite{Aero}. Such methods are typically lightweight and can even operate in a training-free manner, relying solely on reconstruction errors from pre-trained models. However, raw reconstruction errors from a single encoding–decoding pass often provide insufficient separability, as real and generated images can yield numerically similar reconstruction errors. This produces substantial overlap between their error distributions (see the red dashed circle in \Cref{fig:vs}), causing threshold-based decision rules to become unreliable. Beyond these empirical limitations, the theoretical foundations of reconstruction-based detectors remain incomplete, limiting interpretability and making it difficult to understand the principles that govern their behavior.


This limitation suggests that reconstruction behavior should be actively probed rather than passively assessed using a single reconstruction pass, as illustrated in \cref{fig:moti}. Structured image editing provides a principled mechanism for introducing controlled and semantically coherent perturbations~\cite{Nichol22GLIDE,2023instructpix2pix}, enabling systematic examination of reconstruction responses. As shown in \cref{fig:moti}~(c) \& (d), operations such as object insertion (Add), content removal (Erase), and semantic-guided regeneration (SemR) modify localized semantic content or spatial regions while preserving the global layout and overall image identity.
These edits induce meaningful and interpretable perturbations that allow us to quantify how reconstruction behavior changes under well-specified semantic modifications. Notably, \cref{fig:moti}~(e) \& (f) reveals a consistent asymmetric phenomenon: for real images, structured edits tend to draw them closer to the generative manifold, producing reduced reconstruction errors; for generated images, the same edits displace them away from the manifold, leading to increased reconstruction errors. This asymmetric shift constitutes a stable and discriminative signal, motivating a formal investigation into edit-induced reconstruction behavior and providing the basis for the framework presented in the following sections.

Building on this insight, we introduce \textbf{E}dit-\textbf{I}nduced \textbf{R}econstruction \textbf{E}rror \textbf{S}hift (\textbf{EIRES}) , a training-free and model-agnostic framework designed to amplify reconstruction discrepancies between real and generated images through structured editing. 
EIRES incorporates a Multi-Edit (ME) module that applies a sequence of controlled semantic modifications, and the resulting reconstruction error shifts are summarized by the maximum edit-induced reconstruction shift, which provides a far more discriminative signal than raw reconstruction errors alone.
To understand the discriminative effect of structured edits, we analyze reconstruction behavior under a local linear approximation of the decoder. By leveraging a characteristic value of the decoder Jacobian, we obtain a lower bound on the expected reconstruction shift, which explains why real and generated images respond differently to the same semantic perturbations.
Extensive experiments across diverse generative models and open-world settings demonstrate that EIRES consistently improves separability and achieves strong generalization without requiring any additional training or model modification.

Our key contributions are summarized as follows:
\begin{itemize}
    \item We propose EIRES, a training-free and model-agnostic framework that enhances real–generated separability via edit-induced reconstruction error shifts.
    
    
    \item We perform a local geometric analysis using a characteristic value of the decoder Jacobian, deriving a lower bound on the edit-induced reconstruction shift and providing theoretical insight into the behavior of EIRES.
    
    \item 
    EIRES delivers consistent performance across the diverse generative models in GenImage and remains robust on the unbiased subset, even when subjected to post-processing operations such as JPEG compression and cropping.
    

\end{itemize}

%% file: sec/2_Related.tex
\section{Related Work}
\noindent\textbf{Reconstruction-Based Detection of Generated Images.} 
Emerging as a popular strategy for identifying AI-generated images, particularly in the context of diffusion models. These methods rely on the observation that real and generated images exhibit different behaviors under the reconstruction pipeline.
DIRE \cite{Dire} is an early work that introduces reconstruction error as a signal for diffusion model detection, showing that real and generated images exhibit different reconstruction behaviors. Building on this idea, SeDID \cite{SeDID} computes multi-step reconstruction errors to improve discriminative performance. LaRE$^2$ \cite{Lare} further enhances detection accuracy through an error-guided feature refinement module. AERO \cite{Aero} proposes a training-free strategy using the autoencoder of latent diffusion models, and demonstrates competitive results using LPIPS as a perceptual similarity metric. Additionally, HFI \cite{HFI} explores high-frequency sensitivity in autoencoder reconstructions, revealing that frequency can effectively expose distributional discrepancies. 
Unlike existing methods that rely on raw reconstruction errors, we focus on reconstruction dynamics under controlled perturbations. Structured edits amplify latent asymmetries, providing a theoretically grounded signal for detecting generated images.
\vspace{1mm}

\noindent\textbf{Latent Space Geometry and Jacobian Analysis.}
The geometry of latent space fundamentally shapes the behavior of generative models. Existing studies on VAEs ~\cite{kingma2013auto} and GANs~\cite{goodfellow2014generative} have shown that latent structure affects interpolation smoothness, generation fidelity, and reconstruction accuracy \cite{arvanitidis2018latent, choi2025analyzing}. Decoder sensitivity to latent perturbations is often analyzed through its Jacobian matrix \cite{sokolic2017robust, hoffman2019robust}. For Latent Diffusion Models (LDMs), recent work has focused mainly on training objectives and generation quality, leaving their reconstruction behavior less understood. AERO \cite{Aero} reported that generated images exhibit much lower reconstruction errors than real images, yet this phenomenon lacks theoretical support. Thus, we leverage the singular value spectra of decoder Jacobians in LDMs to derive a theoretical lower bound on reconstruction error, providing a principled explanation for the gap between real and generated images.

\vspace{1mm}
\noindent\textbf{Editing Techniques in Generative Models.}
Structured editing with diffusion models has become a powerful tool for controllable image manipulation. Recent methods such as InstructPix2Pix \cite{2023instructpix2pix} and Prompt-to-Prompt \cite{hertz2023prompt} enable text-guided modifications, while object insertion and masked patch regeneration are supported by modern generative models \cite{tewel2025addit, flux} and commercial platforms such as Doubao and Baidu. Existing work focuses on improving editing quality, whereas we treat editing as a structured perturbation to expose latent sensitivity differences between real and generated images. This ``editing-as-probing" perspective is novel in the context of generative image detection.

%% file: sec/3_Method.tex
\begin{figure*}[t!]
\begin{center}
\includegraphics[width=0.9\linewidth]{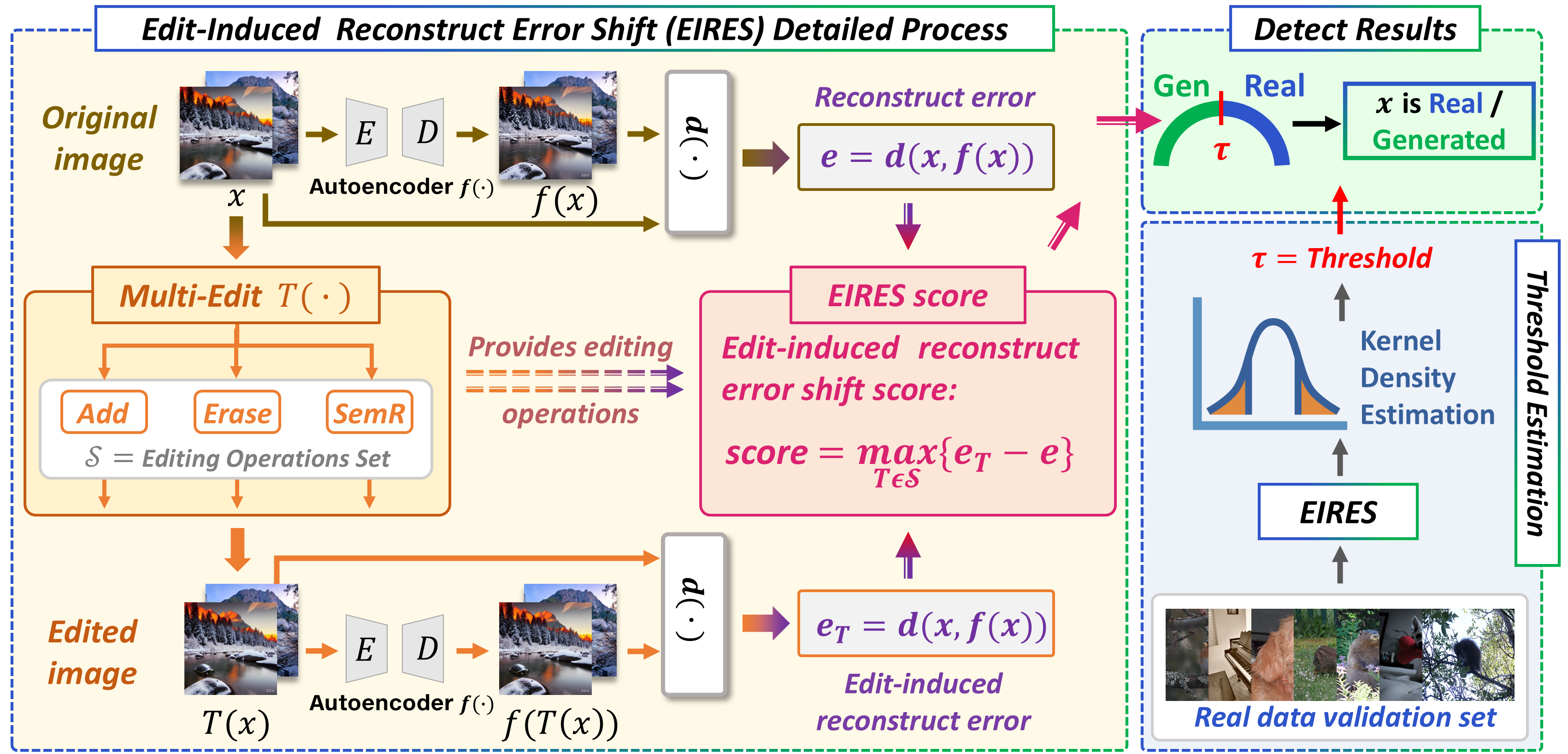}
\end{center}
\vspace{-4mm}
\caption{\textbf{Overview of EIRES.} We aim to establish a boundary between real and generated images without requiring training, enabling the use of threshold-based methods to distinguish between them. The original image is first processed through an autoencoder to compute its reconstruction error. The image is then modified using a \textbf{M}ulti- \textbf{E}dit module (ME), which applies a series of structured edits, including Add, Erase, and SemR. Editing operations are applied and the maximum change in reconstruction error before and after the edits is identified. This maximum error defines the \textbf{E}dit-\textbf{I}nduced \textbf{R}econstruction \textbf{E}rror \textbf{S}hift score (EIRES). The image is classified as real or generated by comparing the EIRES score to a threshold determined through a real data validation set.
}
\label{fig:overview}
\end{figure*}

\section{Motivation}
Reconstruction-based detection becomes increasingly unreliable as modern generative models approach near-photorealistic quality. As shown in \Cref{fig:vs}, raw reconstruction errors from a single autoencoder or diffusion pass often provide limited separability, since real and generated images may produce errors with numerically similar errors that lead to substantial overlap in their reconstruction-error distributions. This limitation suggests that reconstruction should be actively probed rather than passively evaluated.

Structured image editing provides a principled way to introduce controlled and semantically meaningful perturbations. Operations such as object insertion (Add), content removal (Erase), and semantic-guided regeneration (SemR) modify local regions while preserving the overall scene structure, enabling a consistent examination of how an image responds to perturbation and subsequent reconstruction. \Cref{fig:moti}~(c) \& (d) illustrate how such edits produce informative deviations that go beyond random noise or adversarial disturbances.
As we analyze these reconstruction responses under structured edits, a consistent empirical pattern emerges. As shown in \Cref{fig:moti} (e) \& (f), real images tend to yield lower reconstruction error after a mild semantic edit, whereas generated images tend to produce higher error. This opposite shift produces a substantially stronger distinction between real and generated images than raw reconstruction alone.

This behavior aligns with a manifold-based interpretation~\citep{Aero}. Real images often lie outside the autoencoder’s learned generative manifold, and semantic edits move them toward regions that are reconstructed more accurately. Generated images lie on or near the manifold, and edits perturb them away from these coherent regions, leading to degraded reconstruction. The resulting divergence in error change for real and generated images provides a reliable and separable cue. These observations motivate the use of edit-induced reconstruction error shift as a detection signal, which forms the foundation of the EIRES framework discussed in the next section.

%% file: sec/4_Theory.tex
\section{Methodology}

\subsection{Preliminaries}
We consider an image space $\mathbb{R}^{H\times W\times 3}$ and a 
reconstruction model consisting of an encoder–decoder pair,
\begin{align}
    E:\mathbb{R}^{H\times W\times 3}\!\to\!\mathbb{R}^{d}, \qquad
    D:\mathbb{R}^{d}\!\to\!\mathbb{R}^{H\times W\times 3}.
\end{align}
Given a latent prior $p_z$ over $\mathbb{R}^{d}$, the decoder induces a 
reconstruction manifold
\begin{align}
    \mathcal{M} = \{D(z) \mid z \sim \mathrm{supp}(p_z)\}
    \subset \mathbb{R}^{H\times W\times 3}.
\end{align}
This manifold represents the set of images that the reconstruction model can 
faithfully reproduce.  
Real and generated images may differ in their geometric relation to 
$\mathcal{M}$, an aspect further analyzed in \Cref{ch:4.3}.  
Notation used throughout the paper is summarized in 
~\Cref{tab:symbol}.


\begin{table}[t]
\centering
\begin{tabular}{@{}ll@{}}
\toprule
\textbf{Symbol} & \textbf{Meaning} \\ \midrule
$\mathcal{U}$ & Tubular neighbourhood of $\mathcal{M}$\\
$x_{r}$ & Query image, off-manifold $\mathcal M$ \\
$\tilde{x}_{r}$ & Nearest point of $x_{r}$ on $\mathcal{M}$, i.e., $\tilde{x}_{r}=\mathcal P(x_{r})$ \\
$T_{\tilde{x}_{r}}\mathcal{M}$ & Tangent space of $\mathcal{M}$ at $\tilde{x}_{r}$ \\
$\varepsilon_{\perp}$ & Normal deviation $\varepsilon_{\perp}=x_{r}-\tilde{x}_{r}$, $\varepsilon_{\perp}\!\perp\!T_{\tilde{x}_{r}}\mathcal{M}$ \\
$z^{\ast}$ & Latent code of $\tilde{x}_{r}$: $z^{\ast}=E(\tilde{x}_{r})$ \\
$J_{E}(x)$ & Jacobian of $E$ at $x$, a $d\times (3HW)$ matrix \\
$J_{D}(z)$ & Jacobian of $D$ at $z$, a $(3HW)\times d$ matrix \\
$\sigma_{\min}(\cdot)$ & Minimum singular value\\
$\sigma_{\max}(\cdot)$ & Maximum singular value \\
$\kappa_{D}$ & condition number $ \kappa_{D}={\sigma_{\max}\!\bigl(J_{D}\bigr)}/{\sigma_{\min}\!\bigl(J_{D}\bigr)}$ \\ 
\bottomrule
\end{tabular}
\caption{Symbols and notation used throughout the paper}
\label{tab:symbol}
\vspace{-1em}
\end{table}

\subsection{Edit-Induced Reconstruction Error Shift}
\Cref{fig:overview} provides an overview of the EIRES pipeline. Given a reconstruction model $f(x)=D(E(x))$ and a  metric $d~(\cdot~,~\cdot)$, 
the raw reconstruction error of an image $x$ is
\vspace{-0.3em}
\begin{align}
    e(x)=d~(f(x),~x).
    \vspace{-0.3em}
\end{align}
Following common practice in reconstruction-based detection~\cite{Aero,HFI}, we use 
LPIPS\textsubscript{v2} as the default distance metric due to its strong 
perceptual sensitivity. As shown in \Cref{ch:5.4}, EIRES remains robust across 
a wide range of alternative distance measures.

\noindent\textbf{Structured Edits.}
As illustrated in the Multi-Edit module of \Cref{fig:overview}, we apply a set 
of structured editing operators 
$\mathcal{S}=\{\mathcal{T}_1,\dots,\mathcal{T}_K\}$.
Each operator produces an edited variant
$x_{\mathcal{T}_k}=\mathcal{T}_k(x)$
that modifies localized regions while keeping the overall semantics intact.

\vspace{0.3em}
\noindent\textbf{Edit-Induced Error Shift.}
For each edit, we measure how reconstruction responds to the perturbation:
\begin{align}
    \Delta_k(x)=e(x_{\mathcal{T}_k})-e(x).
    \vspace{-0.2em}
\end{align}
\noindent\textbf{EIRES Score.}
To capture the most discriminative edit direction, EIRES aggregates the 
maximum shift:
\vspace{-0.5em}
\begin{align}
    \mathrm{score}(x)
    =\max_{k=1,\dots,K}\Delta_k(x).
    \vspace{-0.5em}
\end{align}
This produces a single scalar summarizing the strongest reconstruction change.  
A threshold $\tau$ is calibrated on a small set of real validation images 
(right panel of \Cref{fig:overview}) and then applied for binary classification:
\begin{align}
    &\mathrm{score}(x)>\tau \;\Rightarrow\; x\text{ is real},\\
    &\mathrm{score}(x)\le\tau \;\Rightarrow\; x\text{ is generated}.
    \vspace{-0.5em}
\end{align}
This shift-based formulation probes how reconstruction behaves under 
semantic perturbations, providing a far more discriminative signal than raw 
reconstruction error alone.








\subsection{Geometric Analysis of Reconstruction Shift}
\label{ch:4.3}
This section analyzes the reconstruction behavior of real and generated images and shows how structured edits induce different shifts, forming the basis of the EIRES.

\begin{figure}[t]
\begin{center}
\includegraphics[width=1\linewidth]{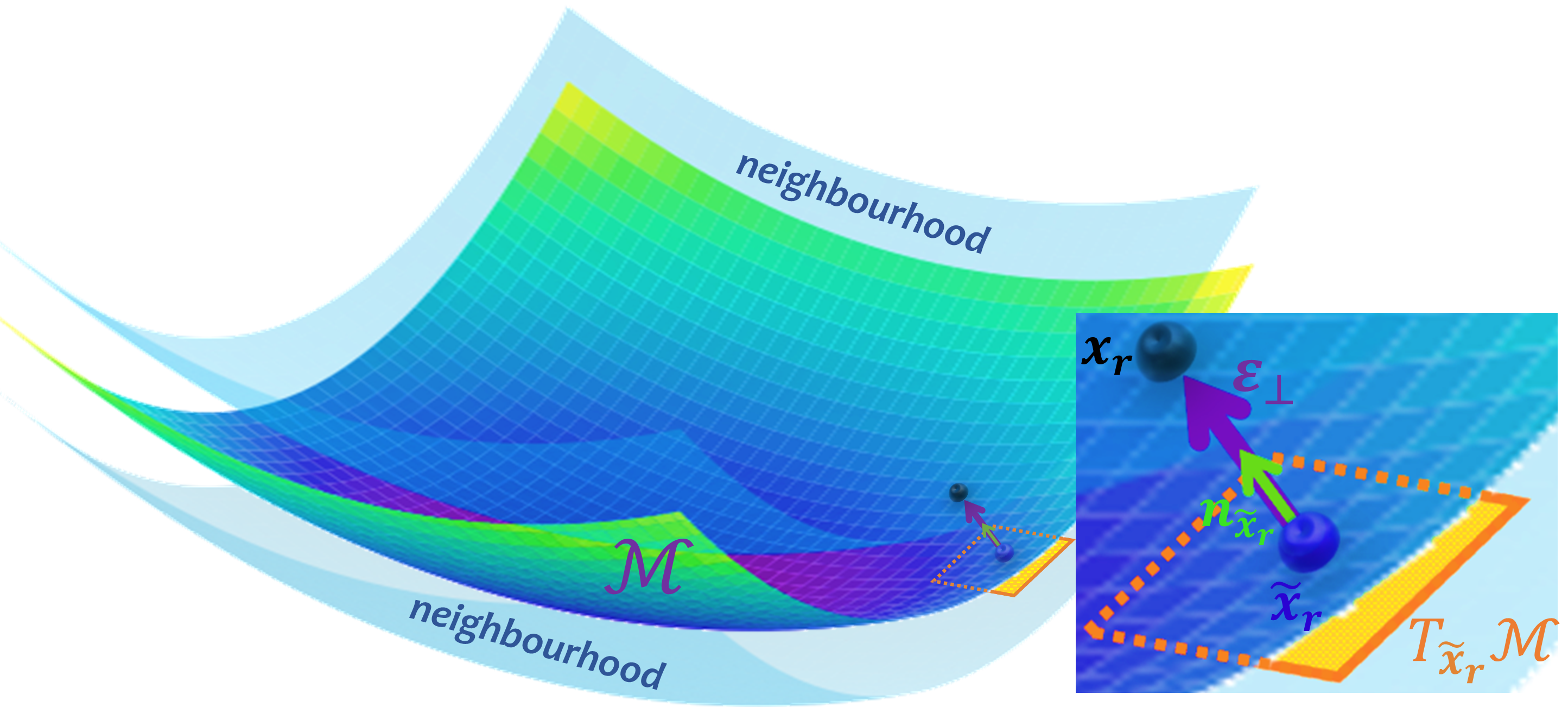}
\end{center}
\vspace{-2mm}
\caption{Geometric interpretation of off-manifold reconstruction.  
A real image $x_r$ is projected onto the reconstruction manifold at 
$\tilde{x}_r$, producing a normal residual $\varepsilon_\perp$ that is 
orthogonal to the local tangent space $T_{\tilde{x}_r}\mathcal{M}$.  
This residual is unavoidable for off-manifold inputs and induces a 
lower-bounded reconstruction error determined by the decoder Jacobian. }
\vspace{-1em}
\label{fig:pro}
\end{figure}

\vspace{0.3em}
\noindent\textbf{Visualization Analysis of Reconstruction Error.} 
A key insight behind EIRES is that 
real images typically lie off the learned manifold, while generated images are located on or near it~\cite{Aero,HFI,Dire}. We provide a geometric illustration in~\Cref{fig:pro} to highlight this distinction. A real image $x_r$, off the manifold, has a unique nearest-point projection $\tilde{x}_r$ onto the manifold, producing a normal residual $\varepsilon_\perp = x_r - \tilde{x}_r$ orthogonal to the tangent space at $\tilde{x}_r$. Since the autoencoder cannot correct this deviation, the reconstruction error is inherently bounded by the local decoder Jacobian. In contrast, generated images, lying on or near the manifold, exhibit minimal residual and reconstruction error.


\vspace{0.3em}
\noindent\textbf{Lower Bound of Reconstruction Error.}  
We now formally define the raw reconstruction errors, marking the first geometric formulation of the lower bound. Unlike existing work focused on empirical observations~\cite{Aero,HFI,Dire}, our approach provides a formal, geometry-based definition.
\vspace{-0.5em}
\begin{proposition}
\label{prop:lower_noedit}
Let $E$ and $D$ be the encoder and decoder of a reconstruction model. For any image $x$ with projection $\tilde{x} \in \mathcal{M}$ and normal deviation $\varepsilon_\perp = x - \tilde{x}$, the reconstruction error satisfies:
\begin{align}
\|x - D(E(x))\|_2 \geq \sqrt{1 + \kappa_D^{-2}} \|\varepsilon_\perp\|_2 + o(\|\varepsilon_\perp\|_2^2),
\end{align}
where $\kappa_D$ is the local condition number of the decoder Jacobian $J_D$ at $\tilde{x}$.
\vspace{-0.5em}
\end{proposition}
The complete proof is provided in the appendix. This formulation establishes the theoretical foundation for understanding the reconstruction error shift induced by structured edits in the following sections.

\vspace{0.3em}
\noindent\textbf{Lower Bound under Edit Perturbation.}
The following proposition formalizes the behavior when  structured edits perturb the input.
\vspace{-0.5em}
\begin{proposition}
\label{prop:lower_edit}
Under the same conditions as Proposition~\ref{prop:lower_noedit}, let $x_{\mathcal{T}} = x + \delta$ denote a structured edit with $\|\delta\|$ sufficiently small.  
Let $\tilde{x}_{\mathcal{T}}$ be its manifold projection and $\varepsilon_{\perp}^{(\mathcal{T})}=x_{\mathcal{T}} - \tilde{x}_{\mathcal{T}}$ the new normal deviation.  
Then:
\begin{align}
    \|x_{\mathcal{T}} - D(E(x_{\mathcal{T}}))\|_2
\ge
\sqrt{1+\kappa_D^{-2}}\|\varepsilon_{\perp}^{\mathcal{T}}\|_2+o(\|\varepsilon_{\perp}^{\mathcal{T}}\|_2^2).\nonumber
\end{align}
Moreover, the reconstruction error shift satisfies
\begin{align}
    \Delta(x)
    =e(x_{\mathcal{T}})-e(x)
    \propto
    \|\varepsilon_{\perp}^{(\mathcal{T})}\|_2 - \|\varepsilon_{\perp}\|_2.
    \vspace{-0.5em}
\end{align}
\end{proposition}
The complete proof is provided in the appendix. Structured edits typically move a real image closer to the manifold, reducing the normal deviation $\|\varepsilon_\perp\|$ and lowering the reconstruction error, leading to a positive shift $\Delta(x)$. In contrast, generated images are more susceptible to larger errors because perturbations induce non-linear shifts in their latent space, increasing the normal deviation $\varepsilon_\perp$ and the reconstruction error.

\vspace{0.3em}
\noindent\textbf{Bridging the Metric Discrepancy.}
Although the theoretical analysis is derived under the Euclidean metric, the practical implementation uses LPIPS~\cite{zhang2018unreasonable}. This is consistent with the theory because LPIPS is locally equivalent to $\ell_2$. As a result, the geometric behavior predicted by the analysis remains valid, and LPIPS offers a perceptual metric that preserves these trends.

%% file: sec/5_Experiment.tex
\begin{table*}[!ht]
\setlength{\tabcolsep}{3.2mm}
\centering
\small 
\begin{tabular}{cccccccccc}
\toprule            
Method  & ADM & BigGAN & GLIDE & Mid & SDv1.4 & SDv1.5 & VQDM & Wukong & Mean \\ 
\midrule 
\multicolumn{10}{c}{Training-based Detection Methods (ACC $\uparrow$)} \\ 
\midrule
CNNSpot \cite{wang2020cnn}  &   0.570 & 0.566 & 0.571 & 0.582 & 0.703 & 0.702 & 0.567 & 0.677 & 0.617\\
Spec \cite{zhang2019detecting}&   0.579 & 0.643 & 0.654 & 0.567 & 0.724 & 0.723 & 0.617 & 0.703 & 0.651\\
F3Net \cite{qian2020thinking} &   0.665 & 0.565 & 0.578 & 0.551 & 0.731 & 0.731 & 0.621 & 0.723 & 0.646\\
GramNet \cite{liu2020global}&   0.587 & 0.612 & 0.653 & 0.581 & 0.728 & 0.727 & 0.578 & 0.713 & 0.647\\
DIRE \cite{wang2023dire}  &   0.619 & 0.567 & 0.691 & 0.650 & 0.737 & 0.737 & 0.634 & 0.743 & 0.672\\
LaRE$^2$ \cite{luo2024lare}  &   0.667& 0.740 & 0.813 & 0.664 & 0.873 & 0.871 & \textbf{0.844} & 0.855& 0.791\\
\midrule
\multicolumn{10}{c}{Training-free Detection Methods (ACC $\uparrow$)} \\ 
\midrule
RIGID \cite{he2024rigid}      & 0.514 & 0.530 & 0.459 & \textbf{0.941} & 0.870 & 0.872 & 0.522 & 0.878 & 0.698\\
MIB \cite{brokman2025manifold} &   0.573 & 0.776 & 0.871 & 0.555 & 0.620 & 0.630 & \underline{0.769} & 0.654 & 0.681\\
AERO~\citep{Aero}  & \underline{0.707} & 0.962 & 0.957 & 0.678 & 0.940 & 0.933 & 0.566 & 0.881 & 0.828 \\
HFI~\cite{HFI}  & 0.695 & \underline{0.974} & \textbf{0.967} & 0.637 & \textbf{0.975} &\textbf{ 0.975} & 0.549 &\underline{0.914}&\underline{ 0.835} \\
EIRES(ours) & \textbf{0.751} & \textbf{0.998} & \underline{0.959} & \underline{0.720} & \underline{0.972} & \underline{0.974} & 0.699 & \textbf{0.935} & \textbf{0.876}\\
\midrule
\multicolumn{10}{c}{Training-free Detection Methods (AUROC $\uparrow$)} \\ 
\midrule
RIGID \cite{he2024rigid}     & 0.572 & 0.530 & 0.539 & \textbf{0.989}& 0.955 & 0.954 & 0.485 & 0.947 & 0.746\\
MIB \cite{brokman2025manifold} &   0.681 & 0.857 & 0.946 & 0.628 & 0.782 & 0.796 & \textbf{0.874} & 0.793 & 0.795\\
AERO~\citep{Aero}   & 0.839 & 0.979 & 0.983 & 0.850 & 0.975 & 0.975 & 0.708& 0.943 & 0.907 \\
HFI~\cite{HFI}  & \underline{0.852} & \textbf{0.990} & \textbf{0.995} & 0.803 & \textbf{0.999} &\textbf{0.999} & 0.763 & \underline{0.974} &\underline{ 0.929} \\
EIRES(ours) & \textbf{0.866} & \underline{0.989} & \underline{0.987} & \underline{0.879} & \underline{0.996} & \underline{0.998} & \underline{0.837} & \textbf{0.979} & \textbf{0.941}\\

\midrule
\multicolumn{10}{c}{Training-free Detection Methods (AP $\uparrow$)} \\ 
\midrule
RIGID \cite{he2024rigid}  & 0.563 & 0.540 & 0.342 & \textbf{0.990} & 0.953 & 0.951 & 0.525 & 0.955 & 0.727\\
MIB \cite{brokman2025manifold} &   0.654 & 0.868 & 0.953 & 0.609 & 0.723 & 0.734 & \textbf{0.878} & 0.766 & 0.773\\
AERO \cite{Aero}   & 0.839 & 0.950 & 0.972 & 0.816 & 0.942 & 0.941 & 0.684 & 0.886 & 0.878\\
HFI~\cite{HFI}  & \underline{0.854} & \textbf{0.987} & \textbf{0.995} & 0.797 & \textbf{0.999} &\textbf{0.999} & 0.716 & \underline{0.972} &\underline{ 0.914} \\
EIRES(ours) & \textbf{0.867} & \underline{0.979} & \underline{0.979} & \underline{0.870} & \underline{0.996} & \underline{0.998} & \underline{0.838}& \textbf{0.979} & \textbf{0.938}\\
\bottomrule
\end{tabular}
\caption{Comparison of EIRES and baseline detection methods across diverse generative models. Detection performance is reported using ACC, AUROC, and AP. Best results are shown in \textbf{bold} and second best are \underline{underlined} (applies to all tables).}   
\vspace{-3mm}
\label{tab:main}
\end{table*}

\section{Experiments}
\subsection{Experimental setups}
\noindent\textbf{Datasets and Metrics.} We evaluate EIRES on the GenImage benchmark~\cite{zhu2023genimage},  which includes real images from ImageNet \cite{imagenet} and generated images from eight representative generators: ADM \citep{ADM}, BigGAN \citep{BigGAN}, GLIDE \citep{Nichol22GLIDE}, Midjourney \citep{22Midjourney}, Stable Diffusion (SD) v1.4/1.5 \citep{Rombach22SD}, VQDM \citep{Gu22VQDM} and Wukong \citep{Gu22Wukong}. We further include the SD v1.4 subset of the Unbiased variant  \cite{Grommelt2024FakeOJ}, which reduces category-level bias by balancing the class distribution between real and generated samples. 
Performance is measured using accuracy (ACC), area under the ROC curve (AUROC) and average precision (AP), following standard practice \cite{he2024rigid, Lare}. 


\noindent\textbf{Baselines.}
We compare EIRES against representative AI-generated image detection methods, covering both training-based and training-free approaches. 
Training-based baselines include CNNSpot~\cite{wang2020cnn}, Spec~\cite{zhang2019detecting}, F3Net~\cite{qian2020thinking}, GramNet~\cite{liu2020global}, DIRE~\cite{wang2023dire} and LaRE$^{2}$~\cite{luo2024lare}. 
Following the GenImage protocol, each model is trained on one generator and evaluated on the remaining ones, with results averaged to obtain a single performance score. Training-free baselines include AERO~\cite{Aero}, which computes perceptual reconstruction error using LPIPS\textsubscript{v2}~\cite{VGG16}, 
RIGID~\cite{he2024rigid}, which leverages the sensitivity of synthetic images to Gaussian perturbations, 
MIB~\cite{brokman2025manifold}, which performs score-function analysis with a pre-trained diffusion model 
and HFI~\cite{HFI}, which examines high-frequency sensitivity in autoencoder reconstructions to reveal distribution-level discrepancies. 
These methods represent widely adopted training-free detection baselines.

\noindent\textbf{Implementation Details.}  
Images are center-cropped and resized to $512\times512$ resolution. 
Stable Diffusion (SD) v1.5 is used as the default reconstruction model. 
Structured edits are generated using ERNIE-iRAG-Edit (ERNIE)~\citep{baidu2025iragedit}. LPIPS\textsubscript{v2} (VGG-2 layer)~\cite{Aero} is used as the distance metric. We apply three edit types consistent with \Cref{fig:moti}:
\begin{itemize}
    \item \textbf{Add}: regenerates the masked region using prompts such as ``add an object in this area".
    \item \textbf{Erase}: removes the masked content and restores with prompts like ``erase the masked area and fill naturally".
    \item \textbf{SemR}: performs a semantic-preserving regeneration of the full with prompts such as ``reconstruct as original".
\end{itemize}
Additional implementation details and code are provided in the supplementary material.


\subsection{Main results}
As shown in \Cref{tab:main}, EIRES achieves the strongest average performance across all eight generative models and all three metrics. It clearly outperforms all supervised baselines, including LaRE$^{2}$, the best-performing trained detector, despite requiring no generated data and no generator-specific optimization. This highlights a key limitation of supervised approaches: their reliance on generator exposure restricts generalization, whereas EIRES maintains high accuracy even when the test generator is completely unseen. Among training-free methods, EIRES achieves the highest mean ACC, AUROC, and AP, and also exhibits the most stable performance across models. By contrast, methods such as RIGID, MIB, AERO and HFI perform well only on certain generator families but drop sharply on others, revealing sensitivity to noise perturbations, frequency characteristics, or score-function assumptions.

Across nearly all generators and all evaluation metrics, EIRES ranks first or second. Even for challenging generators such as Midjourney and VQDM, where other zero-shot methods degrade substantially, EIRES remains consistently strong. These results indicate that edit-induced reconstruction shifts provide a more universal and generator-agnostic signal than raw reconstruction errors or handcrafted perturbations. Overall, EIRES demonstrates both the best average performance and the most reliable cross-generator generalization among all zero-shot detectors, confirming its robustness and practical applicability across GAN-based, diffusion-based and proprietary generative models.

\subsection{Results on Unbiased Dataset}
We further evaluate EIRES on the unbiased SDv1.4 subset, where resolution, format and compression artifacts are removed following \cite{Grommelt2024FakeOJ}. This setup eliminates spurious cues and enforces a strictly artifact-free comparison between real and generated images.
As shown in Table~\ref{table:sdv14-calibration}, EIRES achieves almost identical AUROC and AP scores before and after calibration under a unified LPIPS metric. The minimal change demonstrates that its discriminative signal is not driven by dataset bias or compression cues.
These results confirm that EIRES captures fundamental differences in edit-induced reconstruction behavior, enabling reliable separation of real and generated images even under strictly unbiased conditions where many detectors collapse.
\vspace{-2mm}


\subsection{Ablation Study}
\label{ch:5.4}

\subsubsection{Influence of Edit Types.} 
\Cref{tab:editing-types} compares four edit settings: Add, Fix, Sem and the Multi Edit (ME) scheme used in EIRES.
All three single edit types perform strongly on SDv1.5 and BigGAN, indicating that EIRES is robust to the choice of structured perturbation under high-quality generators.
VQDM shows larger variation across edits due to its lower visual fidelity.
ME consistently achieves the best performance across all generators, demonstrating that aggregating multiple edits yields a more stable and discriminative detection signal than any single edit alone.
\begin{table}[!t]
\centering
\begin{adjustbox}{width=0.95\columnwidth}
\begin{tabular*}{\linewidth}{@{\extracolsep{\fill}}cccc}
\toprule
 & \multicolumn{3}{c}{SDv1.4} \\ \cmidrule(lr){2-4}
Setting & ACC & AUROC & AP\\ 
\midrule
Default-GenImage~\cite{zhu2023genimage} & 0.972 & 0.996 & 0.996\\
Unbiased-GenImage~\cite{Grommelt2024FakeOJ} & 0.966 & 0.996 & 0.996\\
\bottomrule
\end{tabular*}
\end{adjustbox}
\vspace{-2mm}
\caption{ Performance under default and unbiased settings.}
\vspace{-1em}
\label{table:sdv14-calibration}
\end{table}

\begin{table}[!t]
{
\setlength{\tabcolsep}{0.65mm}
\centering
\small
\begin{tabular}{ccccccccccc}
\toprule
 & \multicolumn{3}{c}{VQDM} & \multicolumn{3}{c}{SDv1.5} & \multicolumn{3}{c}{BigGAN} \\ \cmidrule(lr){2-4}\cmidrule(lr){5-7}\cmidrule(lr){8-10}
  Edit type& ACC & AUC & AP& ACC & AUC & AP & ACC & AUC & AP\\ 
\midrule
Add & 60.4 & 76.3 & 76.0 & \underline{97.4} & \textbf{99.9} & \textbf{99.9} & 96.3 & 98.4& 96.9 \\
Fix & \underline{68.6} & \underline{81.7}& \underline{82.3} &97.3  &\textbf{99.9} & \textbf{99.9} &94.8 & 98.2& 96.8\\
Sem & 55.8 & 66.1 & 65.3 & 97.3 & 99.7 & 99.6 & 94.4 & 98.3 & 97.6\\
ME(ours) & \textbf{69.9} &\textbf{ 83.7} & \textbf{83.8} & \underline{97.4} & \underline{99.8} & \underline{99.8}  & \textbf{99.8}  & \underline{98.8}& \underline{97.9} \\
\bottomrule
\end{tabular}}
\vspace{-2mm}
\caption{Detection performance (\%) across different structured edit types. `AUC' denotes `AUROC'.}
\vspace{-1em}
\label{tab:editing-types}
\end{table}
\begin{table}[!t]
{
\setlength{\tabcolsep}{1.4mm}
\centering
\small
\begin{tabular}{cccccccc}
\toprule
 & \multicolumn{3}{c}{BigGAN} & \multicolumn{3}{c}{SDv1.4} \\ \cmidrule(lr){2-4}\cmidrule(lr){5-7}
  Edit model& ACC & AUROC & AP& ACC & AUROC & AP\\ 
\midrule
ERNIE & \textbf{0.969} & \textbf{0.988}& \textbf{0.979} & \textbf{0.972} & \textbf{0.996} & \textbf{0.996}\\
FLUX.1 & 0.944 & 0.982 & \textbf{0.979}  & 0.896 &  0.969& 0.949 \\
\bottomrule
\end{tabular}
}
\vspace{-2mm}
\caption{Detection performance across different editing models.}
\vspace{-1em}
\label{table:editing-model}
\end{table}
\begin{table}[!t]
{
\setlength{\tabcolsep}{0.7mm}
\centering
\small 
\begin{tabular}{cccccccccc}
\toprule                         
AE& AD & BG & GL & Mid & SD1.4 & SD1.5 & VQ & WK & Mean\\ 
\midrule 
SDv1.5 & \textbf{86.7} & \underline{97.9} & \underline{97.9} & \underline{82.7} & \textbf{99.6} & \textbf{99.8} & \underline{83.8} & \textbf{97.9} & \textbf{93.2}\\
SDv2.1 & 83.5 & 96.7 & 97.4 & 81.3 & 78.0 & 77.6 & 81.3 & 71.8 & 83.4 \\
Kandinsky & \underline{84.8} & \textbf{98.1} & \textbf{98.3} & \textbf{84.7} & 67.5 & 67.1 & \textbf{84.2} & 62.0 &80.8\\
Ensemble & 84.6 & 97.4& 97.7 & \underline{82.7} & \underline{99.4} & \underline{99.4} & 82.4 & \underline{95.1} &\underline{92.3}\\
\bottomrule
\end{tabular}
}
\vspace{-2mm}
\caption{AP (\%) with different autoencoder backbones.}
\label{tab:abla_ae}
\vspace{-1em}
\end{table}

\subsubsection{Influence of Editing Models.}
We compare two editing pipelines: ERNIE-iRAG-Edit, a high-fidelity commercial editor, and FLUX.1, an open-source diffusion-based model. As shown in Table~\ref{table:editing-model}, EIRES achieves consistently strong performance with both editors. The differences on SDv1.4 are small, with ERNIE providing slightly higher accuracy due to its stronger semantic consistency. Overall, EIRES is largely insensitive to the choice of editing model, confirming that edit-induced reconstruction shifts remain stable across editing backbones.

\subsubsection{Influence of Autoencoders.}
\Cref{tab:abla_ae} evaluates EIRES with four reconstruction backbones. SDv1.5 yields the highest mean AP (92.8\%), outperforming SDv2.1 and Kandinsky 2.1 (KD2.1) \citep{razzhigaev2023kandinsky}, both of which exhibit notable drops on multiple generators. The Ensemble strategy~\citep{Aero}  achieves competitive accuracy but requires multiple reconstructions per image, making it significantly less efficient. Overall, SDv1.5 provides the best trade-off between accuracy and computational cost, validating its use as the default reconstructor. Importantly, EIRES maintains strong performance across all autoencoders, indicating that its edit-induced reconstruction shift is robust to reconstruction backbone choice.



\subsubsection{Influence of Distance Metrics.}
As shown in Table~\ref{table:ab_dis}, EIRES performs consistently well across different distance metrics, demonstrating robustness to the distance choice. Among the evaluated metrics, LPIPS\textsubscript{v2}~\cite{zhang2018unreasonable} achieves the highest mean AP, surpassing both other perceptual variants (e.g., LPIPS\textsubscript{V}~\cite{VGG16}, LPIPS\textsubscript{A}~\cite{Alex}) and traditional metrics (e.g., MSE, SSIM, PSNR). Although simpler metrics like MSE and PSNR also yield competitive results, LPIPS\textsubscript{v2} consistently provides the best balance of performance across generators.
These results validate LPIPS\textsubscript{v2} as the most effective distance metric for EIRES, ensuring high detection performance and generalizability.

\begin{table}[t]
{
\setlength{\tabcolsep}{1mm}
\centering
\small
\begin{tabular}{cccccccccc}
\toprule                         
Dist & AD & BG & GL & Mid & SD1.4 & SD1.5 & VQ & WK & Mean\\  
\midrule 
$\text{L}_\text{V}$ & 78.1 & 88.0 & 94.8 & \textbf{84.7} & \textbf{99.6}& \underline{99.7} & \underline{77.2} & \textbf{97.9} & \underline{90.0}\\
$\text{L}_\text{V2}$ & \textbf{86.7} & \textbf{97.9} & \textbf{97.9} & \underline{82.7} & \textbf{99.6}& \textbf{99.8} & \textbf{83.8} & \textbf{97.9}& \textbf{93.2}\\
$\text{L}_\text{A}$ & 75.3 & 82.5 & 94.9 & 82.2 & 96.2 & 96.1 & 58.0 & \underline{94.6} & 85.0\\
$\text{L}_\text{S}$ & 73.3 & 81.0 & 92.4 & 84.5 &\underline{98.9} & 98.9 & 50.3 & \underline{94.6} & 86.0\\
DIS & 62.1 & 69.4 & 80.2 & 68.4 & 93.1 & 92.5 & 43.8 & 91.1 & 75.1\\
$\text{MSE}$ & 80.6 & 96.2 & \underline{96.5} & 74.8 & 96.2 & 96.2 & 68.0 & 94.8 & 87.9\\
$\text{SC}$ & 82.2 & \underline{96.4} & 96.0 & 74.1 & 96.5 & 96.5 & 71.8 & 92.3 & 88.2\\
$\text{SS}$ & \underline{83.9} & 96.2 & 96.0 & 72.8 & 94.6 & 94.5 & 75.3 & 91.3 & 88.1\\
$\text{PSNR}$ & 79.8 & 93.9 & 94.5 & 63.2 & 82.2 & 82.5 & 69.4 & 71.1 & 79.6\\
$\text{MS}$ & 78.7 & 91.9 & 92.0 & 66.9 & 81.5 & 81.9 & 70.9 & 71.9 & 79.5\\
\bottomrule
\end{tabular}
}
\vspace{-2mm}
\caption{AP (\%) measured across different distance functions}
\label{table:ab_dis}
\vspace{-2mm}
\end{table}

\begin{table}[!t]
{
\setlength{\tabcolsep}{0.65mm}
\centering
\small
\begin{tabular}{ccccccccccc}
\toprule
 & \multicolumn{3}{c}{VQDM} & \multicolumn{3}{c}{SDv1.5} & \multicolumn{3}{c}{BigGAN} \\ \cmidrule(lr){2-4}\cmidrule(lr){5-7}\cmidrule(lr){8-10}
  Compute type& ACC & AUC & AP& ACC & AUC & AP & ACC & AUC & AP\\ 
\midrule
Mean & 62.7 &78.9 & 79.3 & \textbf{97.5} & \textbf{99.9} &\textbf{99.9}  & \underline{96.5} &\textbf{99.1}& \textbf{98.3} \\
Min & 56.4 &68.4 & 67.7 & 97.3 & \textbf{99.9}& \textbf{99.9} & 94.4 & 98.5&97.8\\
Max(ours) & \textbf{69.9} &\textbf{ 83.7} & \textbf{83.8} & \underline{97.4} & \underline{99.8} & \underline{99.8}  & \textbf{99.8}  & \underline{98.9}& \underline{97.9} \\
\bottomrule
\end{tabular}}
\vspace{-2mm}
\caption{Detection performance (\%) for each structured edit. ``AUC” denotes AUROC.}
\label{tab:editing-types}
\vspace{-1.5em}
\end{table}

\subsubsection{Influence of the Max-Selection Strategy.}
Table~\ref{tab:editing-types} evaluates the effect of the Max-selection rule in EIRES, which chooses the edit that produces the largest reconstruction error shift among multiple structured edits. This strategy consistently improves performance. It delivers the highest AP on VQDM and achieves strong accuracy on both SDv1.5 and BigGAN, outperforming the mean and min aggregation variants. The gains are most evident on VQDM, where edit responses vary more widely due to lower image quality, suggesting that selecting the strongest edit response yields a more reliable separation. Although Max-selection incurs additional editing cost, EIRES remains competitive even under single-edit settings. Overall, these results confirm that the Max-selection rule provides the most effective and stable aggregation strateg.


\subsection{Robustness to Unseen Perturbations}
\begin{figure}[!t]
\begin{center}
\includegraphics[width=1\linewidth]{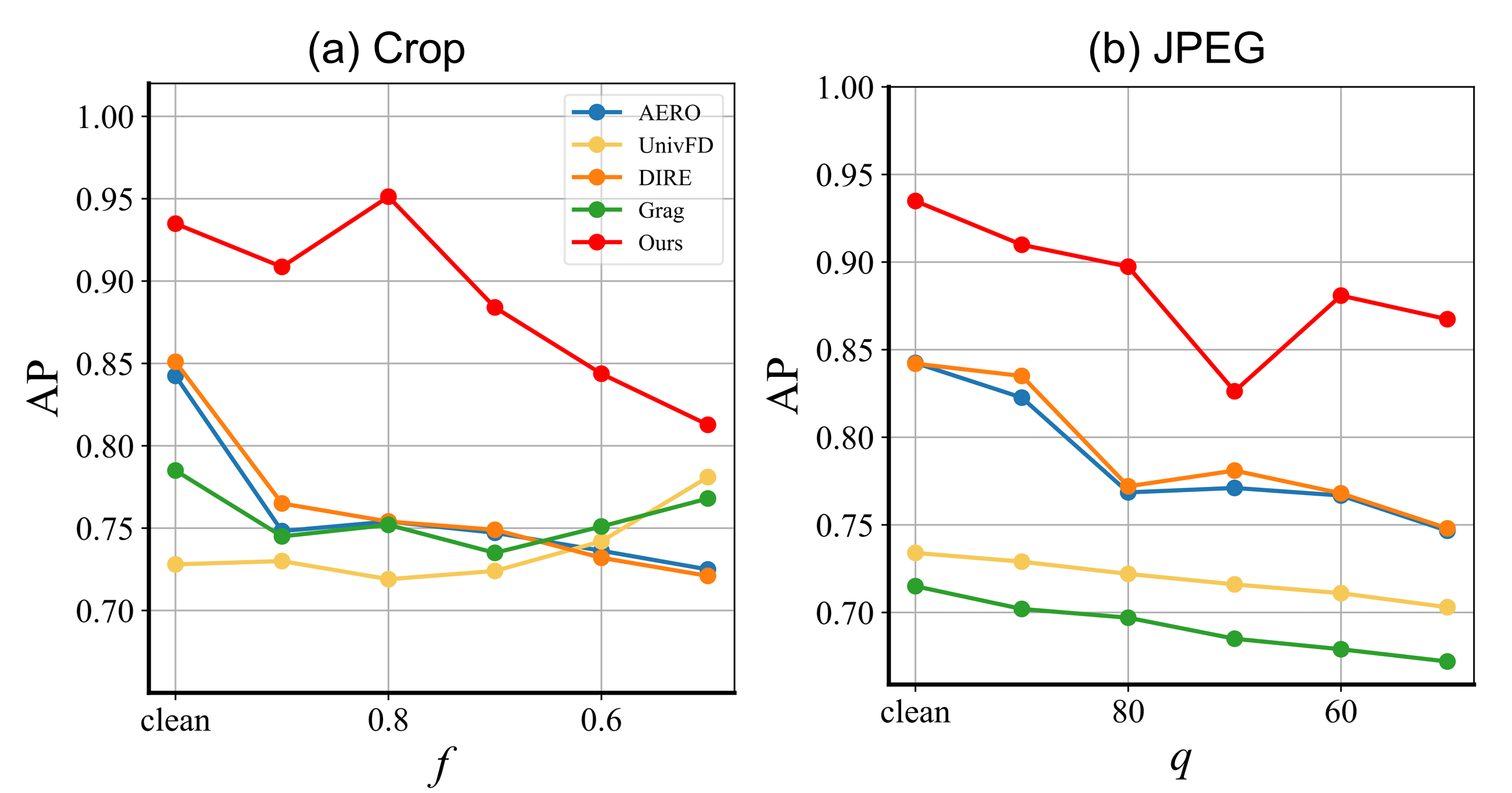}
\end{center}
\vspace{-2em}
\caption{Detection robustness under real-world degradations. \textbf{(a)} AP under different crop ratios $f$. \textbf{(b)} AP under varying JPEG compression quality $q$.}
\label{fig:perturb}
\vspace{-6mm}
\end{figure}

In real-world scenarios, images on social media often undergo post-processing (e.g., cropping and compression), which can degrade detection performance. To assess robustness under such conditions, we simulate two common perturbations: JPEG compression (quality $q$) and center cropping (ratio $f$), followed by resizing, as done in existing works \cite{Aero, HFI}. As shown in \Cref{fig:perturb}, EIRES (red curve) consistently outperforms baselines across various degradation levels. In the cropping case (\Cref{fig:perturb}(a)), EIRES maintains stable AP as $f$ decreases from 1.0 to 0.5, peaking at $f = 0.8$, likely due to better focus on salient regions. In the JPEG compression scenario (\Cref{fig:perturb}(b)), while other methods degrade significantly with reduced image quality, EIRES experiences only a slight drop in AP. Interestingly, a local dip at $q = 70$ may be due to compression artifacts interfering with the editing-induced perturbations, but performance quickly recovers at $q = 60$, demonstrating robustness to signal-level distortions. These results show that EIRES remains robust for clean and post-processed inputs.

%% file: sec/6_Conclusion.tex
\section{Conclusion}
In this paper, we introduced EIRES, a simple and training-free framework for detecting AI-generated images by leveraging a consistent reconstruction asymmetry induced by structural edits. The resulting edit-induced reconstruction-error shift provides a highly discriminative signal, and our theoretical lower bound explains why it enlarges the real–fake margin and enables stable zero-shot thresholding. Extensive experiments show that EIRES is effective, robust to unseen perturbations, and competitive even without auxiliary data. Future work includes exploring broader edit-induced cues, extending EIRES to more complex generative models and multimodal setting and integrating it with practical content authenticity pipelines. 


%% file: sec/X_suppl.tex
\clearpage
\setcounter{page}{1}
\maketitlesupplementary

\section{Summary}
For clarity and completeness, we provide additional details of our method in the supplementary material. The supplementary document is organized as follows:
\begin{itemize}
    \item In \Cref{EIRES}, (i) we provide the pseudocode and an illustrative example of the EIRES detection pipeline, and (ii) additional details of the experimental setup.
    \item In \Cref{visual}, we show visual comparisons between recent raw reconstruction error methods and our proposed EIRES, highlighting error distribution histograms on the GenImage dataset and under JPEG compression and cropping.
    \item In \Cref{proofs}, we provide the complete proofs of \textbf{Proposition~1} and \textbf{Proposition~2} from Section~\textcolor{red}{4.3} of the main paper, covering the lower bound of raw reconstruction error and the lower bound under edit perturbation.

    \item In \Cref{future}, we discuss the limitation of
    our method and outline our future research.
\end{itemize}

\section{Method and Implementation Details}
\label{EIRES}

\begin{figure*}[!h]
\begin{center}
\includegraphics[width=1\linewidth]{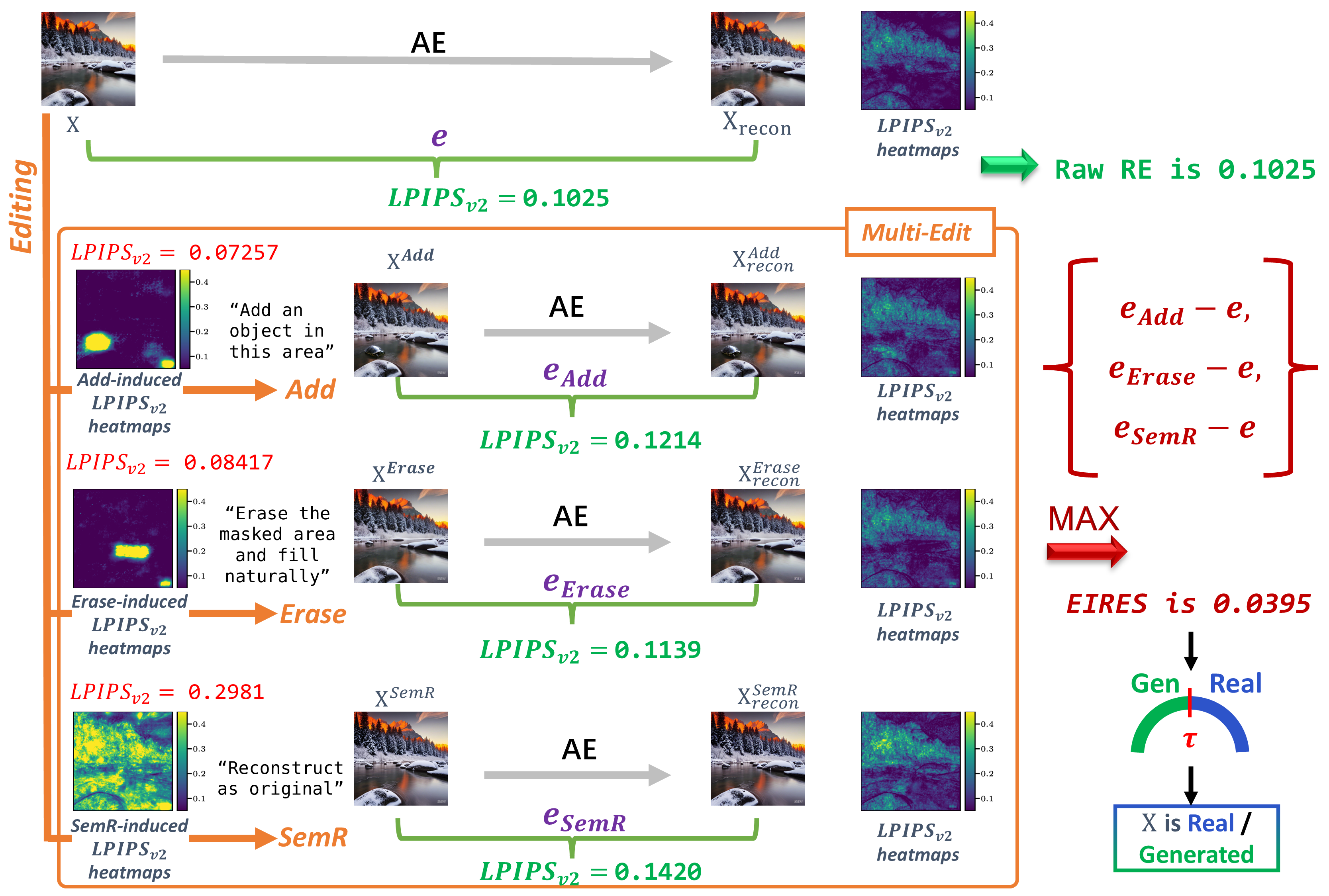}
\end{center}
\caption{Example of our method for detecting an image. On the left side of the figure, we show the LPIPS distance and visual results between the original image and the edited versions induced by structured edits, highlighting the controllable nature of the edits. It is important to note that in the actual detection process, calculating these distances is not necessary. The final detection score, EIRES, is derived from the maximum deviation after applying multiple edits.}
\label{fig:method}
\end{figure*}

\begin{figure*}[!h]
\begin{center}
\includegraphics[width=1\linewidth]{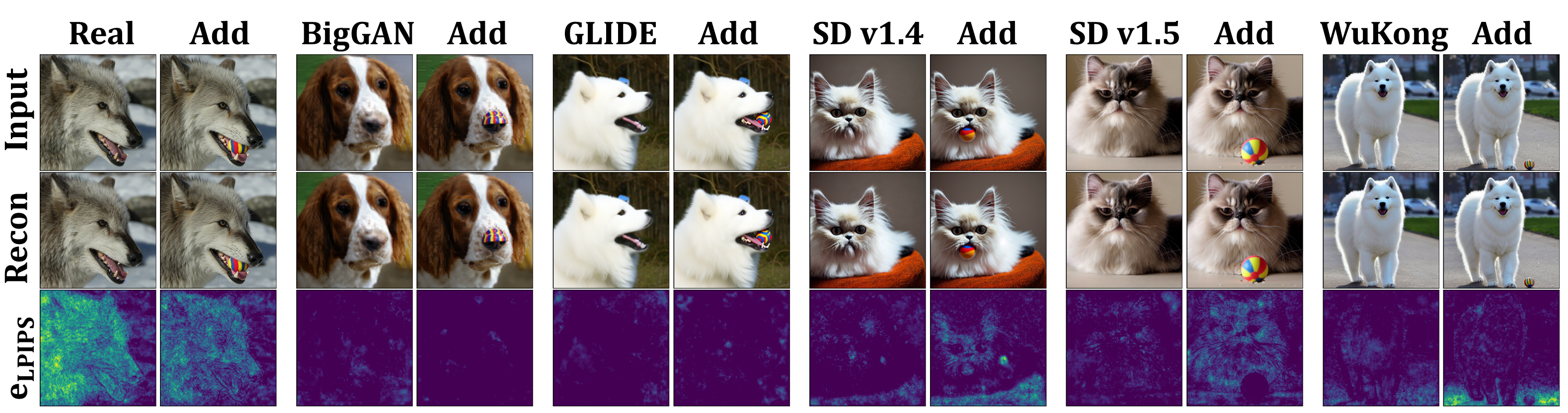}
\end{center}
\caption{Visualization of reconstruction behavior under the Add editing operation across different generative models.
The Add operation is implemented using Add-it~\cite{tewel2025addit} based on FLUX.1, with a circular mask applied to the image and the prompt “Insert a small and brightly colored ball.”
For each model, we show the input image, its autoencoder reconstruction, and the corresponding LPIPS heatmap of reconstruction error.
Real images (leftmost) exhibit noticeably reduced reconstruction error after editing, whereas generated images (BigGAN, GLIDE, SD v1.4, SD v1.5, Wukong) display degraded or unstable reconstructions.
This contrast illustrates the asymmetric edit-induced reconstruction shift that EIRES leverages for distinguishing real from generated images.}
\label{fig:Add}
\end{figure*}

\subsection{EIRES Detection Pipeline}

The EIRES detection pipeline is summarized in the pseudocode provided in \Cref{alg:EIRES}. Given an input image, either real or generated, the method applies a predefined set of structured edits and performs autoencoder-based reconstruction for each edited version. The reconstruction deviations relative to the original image are then computed, and the maximum deviation across all edits is used as the final detection score.
To complement the pseudocode, \Cref{fig:method} offers a visual example of the detection workflow, illustrating how structured edits are generated and how the corresponding reconstruction shifts are measured.

\subsection{Implementation Details}
In our main experiments and ablation studies, we employ the ERNIE-iRAG-Edit model from the Baidu Qianfan platform~\cite{baidu2025iragedit} and report results under the Multi-Edited  strategy. For Add, we apply a circular mask of 75-pixel radius at the image center with the prompt “Add an object in this area”. The Erase operation uses the same mask with the prompt “Erase the masked area and fill naturally”. For SemR, no mask is used and the prompt is “Reconstruct as original”.
For experiments in Section \textcolor{red}{5.3} (Results on Unbiased Dataset) and Section \textcolor{red}{5.5} (Robustness to Unseen Perturbations), which require evaluating performance under multiple crops and JPEG compression levels. To reduce API cost at scale, we switch to the open-source FLUX.1 model~\cite{flux} and apply the simplest Sem editing strategy using the same prompt. As shown in Figure~\ref{fig:Add}, FLUX.1 produces stable and controllable edits comparable to those of commercial editing engines, ensuring reliable evaluation in large-scale experiments.

The implementation code is provided in the supplementary material. All experiments are conducted on a workstation running Ubuntu 20.04.6 LTS with Python 3.10.16 and an NVIDIA RTX A6000 GPU. We implement EIRES using PyTorch 2.1.2 with CUDA 12.1, and employ NumPy 1.26.3 for numerical computation.

Although our experiments involve applying edits to a large number of images, which leads to non-negligible API and computational cost, this overhead comes entirely from the requirements of experimental evaluation rather than from the EIRES method itself. In real-world deployment, EIRES only needs to apply editing operations to the images being checked, and such edits can be generated efficiently. Modern commercial editing engines (e.g., Doubao and Midjourney~\cite{doubao2024,midjourney2024}) can produce high-quality structured edits rapidly and at negligible cost, enabling EIRES to be integrated into content-moderation pipelines, security review workflows, or on-device detection modules with minimal overhead. As a result, EIRES is scalable, hardware-efficient, and practical for large-volume or latency-sensitive applications.



\begin{figure*}[!t]
\begin{center}
\includegraphics[width=1\linewidth]{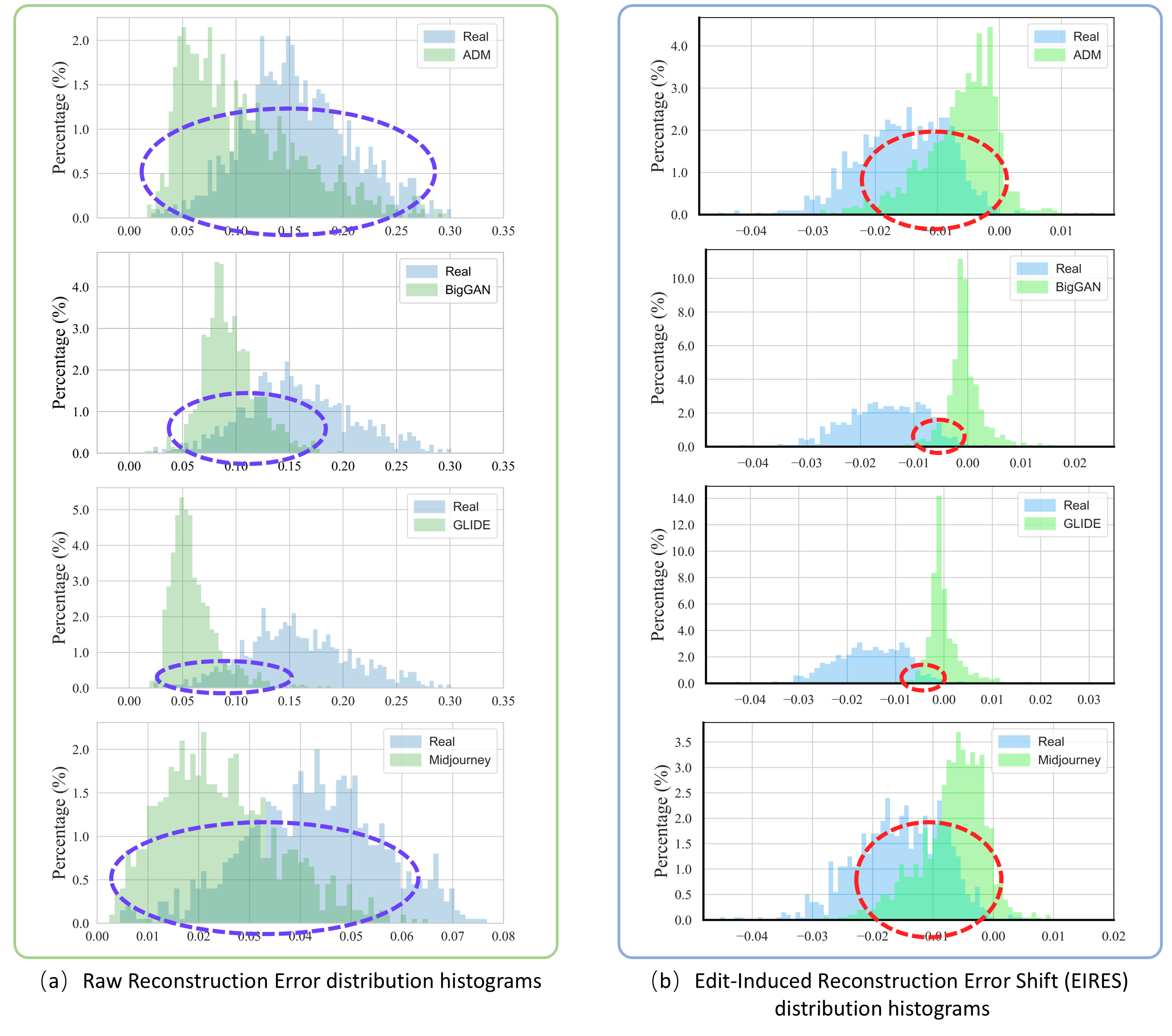}
\end{center}
\vspace{-2em}
\caption{Comparison of score distributions between real and generated images across four generative models: ADM, BigGAN, GLIDE, and Midjourney.
(a) Raw reconstruction error distributions show substantial overlap between real and generated images, as highlighted by the purple dashed circles, indicating limited discriminability.
(b) EIRES score distributions exhibit a much clearer separation, where real and generated images form distinct clusters. The red dashed circles emphasize the regions where EIRES successfully enlarges the separation margin, revealing a significantly more discriminative signal than raw reconstruction error.}
\label{fig:3}
\vspace{-2mm}
\end{figure*}

\begin{figure*}[!t]
\begin{center}
\includegraphics[width=1\linewidth]{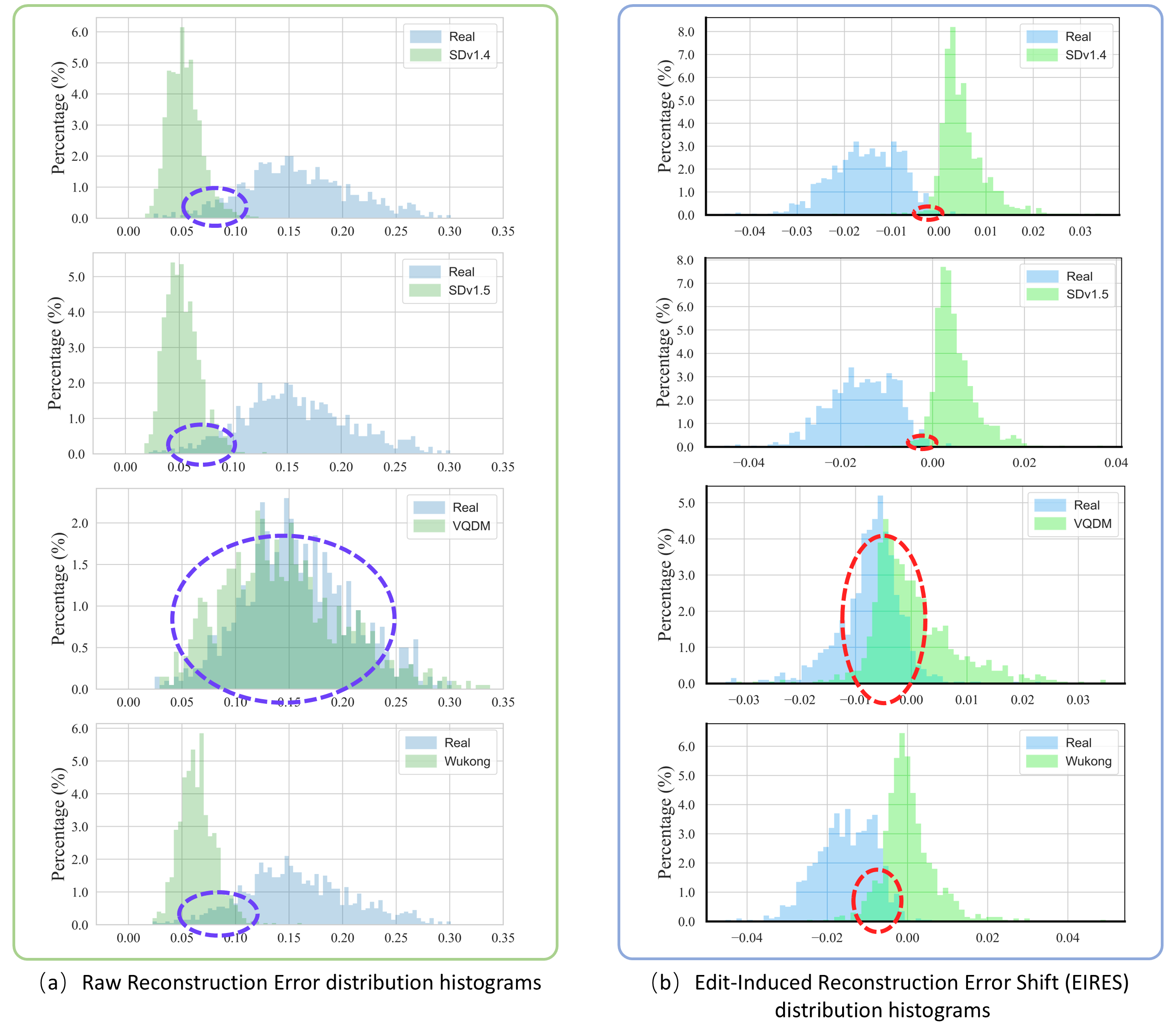}
\end{center}
\caption{Comparison of score distributions between real and generated images across four generative models: SD v1.4, SD v1.5, VQDM, and Wukong.
(a) Raw reconstruction error distributions show substantial overlap between real and generated images, as highlighted by the purple dashed circles, indicating limited discriminability.
(b) EIRES score distributions exhibit a much clearer separation, where real and generated images form distinct clusters. The red dashed circles emphasize the regions where EIRES successfully enlarges the separation margin, revealing a significantly more discriminative signal than raw reconstruction error.}
\label{fig:4}
\vspace{-2mm}
\end{figure*}

\begin{figure*}[!t]
\begin{center}
\includegraphics[width=1\linewidth]{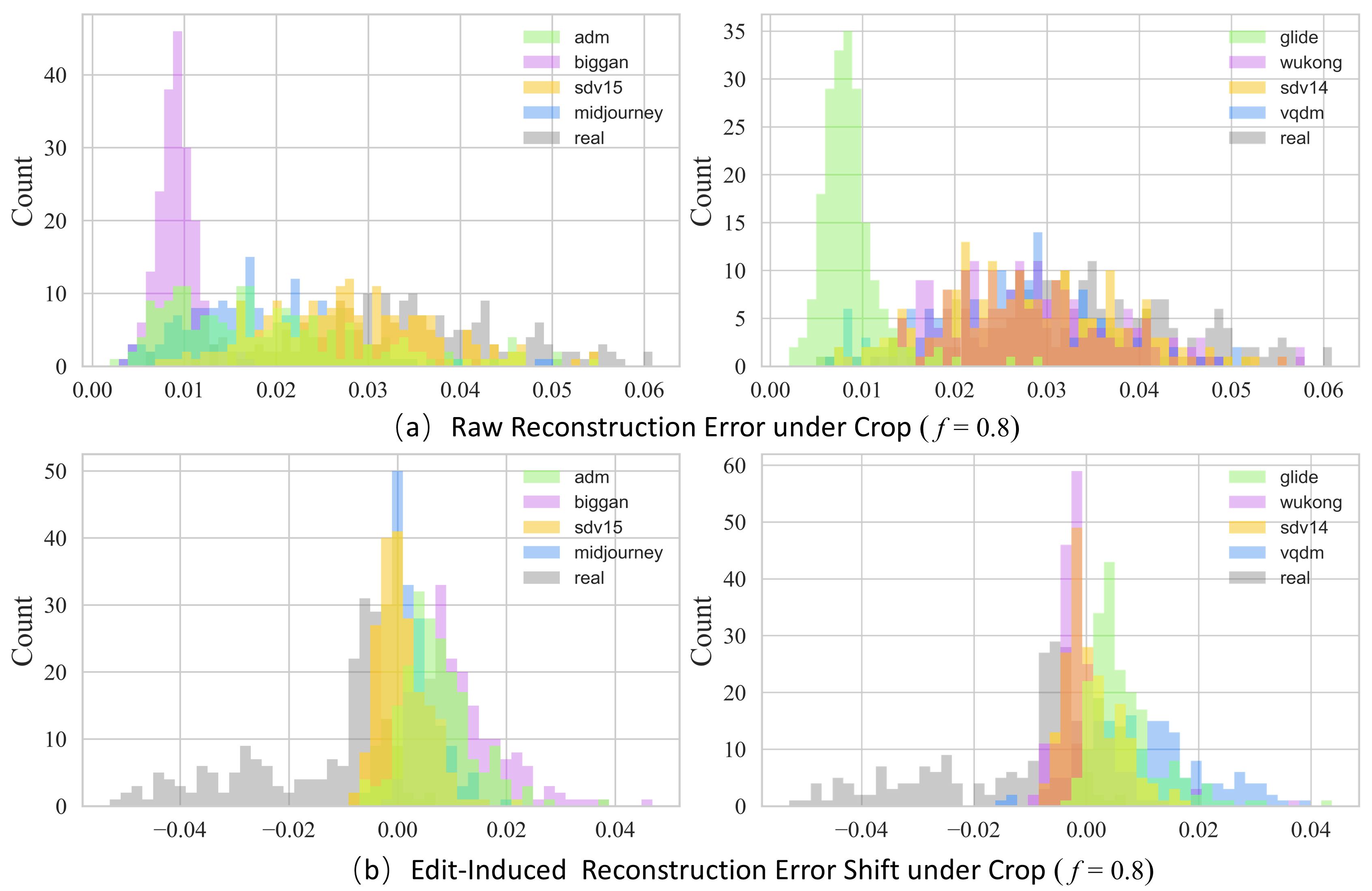}
\end{center}
\caption{Distribution of detection scores under center-crop perturbation. We compare raw reconstruction error (top row) with the EIRES edit-induced reconstruction error shift (bottom row) across eight generative sources (ADM, BigGAN, VQDM, SD v1.4, SD v1.5, GLIDE, Wukong, Midjourney) under a center crop with ratio $f = 0.8$. Raw reconstruction errors exhibit substantial overlap between real and generated images, while EIRES produces more separable distributions, demonstrating improved robustness to input degradation.}
\label{fig:crop}
\vspace{-2mm}
\end{figure*}

\begin{figure*}
\begin{center}
\includegraphics[width=1\linewidth]{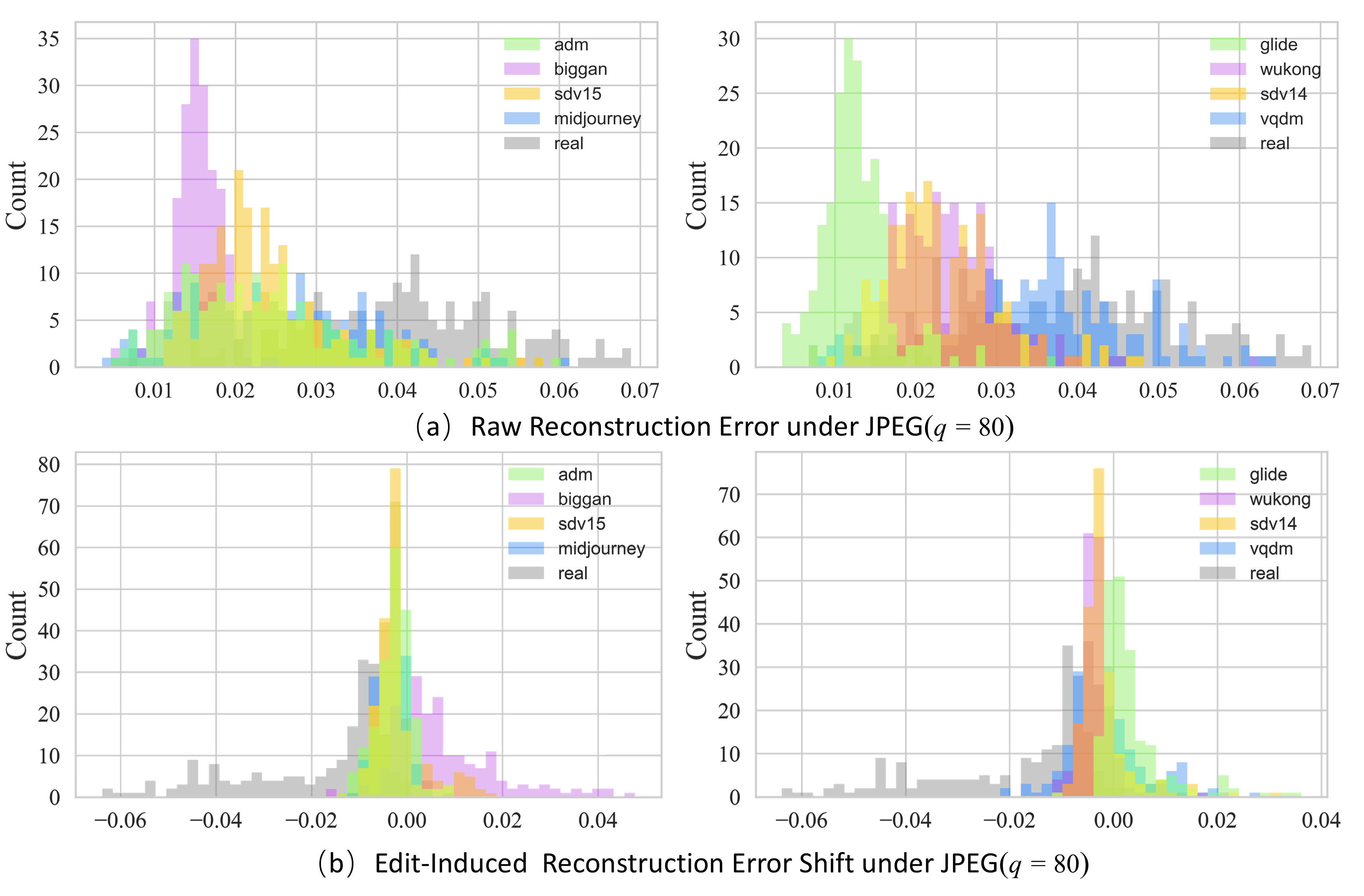}
\end{center}
\caption{Distribution of detection scores under JPEG compression.
We compare raw reconstruction error (top row) with the EIRES edit-induced reconstruction error shift (bottom row) for real images and generated images from eight sources under JPEG compression with quality factor $q = 80$. Across all generative models, EIRES produces significantly more separable real–fake score distributions than raw reconstruction error, demonstrating improved robustness to compression-induced degradation.}
\label{fig:jpeg}
\vspace{-2mm}
\end{figure*}

\begin{algorithm}[t]
\caption{EIRES Detection Pipeline}
\label{alg:EIRES}
\textbf{Input}: input image or a set of images $x$\\
\textbf{Output}: detection result \texttt{Label}\\
\textbf{Notation:}
\begin{itemize}
  \item $\mathcal{S}$: the set of structured editing operations
  \item $T(\cdot)$: structured editing operation
  \item $f(\cdot)$: pre-trained AutoEncoder
  \item $d(\cdot, \cdot)$: LPIPS perceptual distance metric
\end{itemize}
\begin{algorithmic}[1]
\STATE $\tilde{x} \leftarrow T(x),\quad T \in \mathcal{S}$
\STATE $e_{\text{pre}} \leftarrow d(x, f(x))$
\STATE $e_{\text{post}} \leftarrow d(\tilde{x}, f(\tilde{x}))$
\STATE $\triangle e^{*} \leftarrow \max_{T \in \mathcal{S}} \{ e_{\text{post}} - e_{\text{pre}} \}$
\IF{$\triangle e^{*} < \tau$}
    \STATE \texttt{Label} $\leftarrow$ real
\ELSE
    \STATE \texttt{Label} $\leftarrow$ fake
\ENDIF
\STATE \textbf{return} \texttt{Label}
\end{algorithmic}
\end{algorithm}

\section{Additional Visual Comparisons}
\label{visual}
\subsection{Edit-Induced Reconstruction Behavior}
To further validate our observations, we adopt the editing strategy from the Add-it \citep{tewel2025addit} framework to visualize the reconstruction behavior. As shown in \Cref{fig:Add}, real images show high reconstruction error before editing, especially around high-frequency textures, but this error decreases after Add editing. In contrast, generated images exhibit increased error post-editing. This indicates that real images are more responsive to structural changes, while generated ones align with the reconstruction manifold only in their original form.
Interestingly, we observe that the impact of editing is not confined to the masked region. Even when the mask is very small, the reconstruction error changes across the entire image. This indicates that structured editing affects the global latent representation, altering the decoder’s behavior beyond the local region. Such non-local effects support our hypothesis that dynamic reconstruction error captures broader perturbation responses, making it a more reliable signal for distinguishing real from generated images.

\subsection{Distribution Comparative Analysis}
As shown in \Cref{fig:3} and \Cref{fig:4}, we visualize the distribution of detection scores for real and generated images, where the generated images come from eight representative sources. Across all settings, EIRES consistently yields more separable distributions between real and generated samples. For instance, in the Midjourney and GLIDE settings, the real and fake distributions under raw reconstruction error exhibit significant overlap, making accurate detection difficult. In contrast, EIRES pulls the two distributions further apart, enabling clearer distinction even without any model-specific training.
This advantage is especially valuable in a training-free setting, where the detector cannot rely on thresholds learned from labeled data. In many practical scenarios, labeled real or generated images may be unavailable or unreliable, especially when new generative models emerge rapidly. A method that can generalize without model-specific supervision becomes essential for scalable and robust deployment. By leveraging dynamic responses induced by structured edits, EIRES enhances the discriminative power of raw reconstruction error. Instead of passively measuring a single raw reconstruction error, EIRES actively perturbs the input through semantic edits, allowing the model to reveal how well the reconstruction aligns with the underlying manifold. This dynamic behavior not only improves separability but also mitigates overfitting to specific generators or datasets. As a result, EIRES exhibits strong generalization across diverse generative sources, making it suitable for open-world detection where the source of generated content is unknown or continuously evolving.


\subsection{Distribution Comparison under Perturbations.}
\Cref{fig:crop} and \Cref{fig:jpeg} compare the detection-score distributions produced by EIRES and by raw reconstruction error under two common degradations: center cropping   and JPEG compression. We use center crop with ratio $f = 0.8$ and JPEG compression with 
$q = 80$ because these settings are standard and commonly encountered in real-world image sharing. A crop of 0.8 preserves most semantics while introducing meaningful structural changes, and JPEG quality 80 reflects typical platform compression. These moderate, realistic perturbations provide a fair and reproducible way to evaluate robustness. Across all generative models, raw reconstruction error distributions show large real–fake overlap after degradation, which collapses the margin needed for reliable thresholding. In contrast, EIRES maintains a clear separation between real and generated images even when inputs are cropped or heavily compressed. This indicates that the edit-induced reconstruction shift is remarkably stable under structural distortion and signal-level degradation. Moreover, EIRES continues to amplify intrinsic manifold differences regardless of the generator or perturbation, reinforcing the benefit of dynamic reconstruction behavior over static, single-pass error measures.
These results further highlight EIRES as a practical and perturbation-robust detection mechanism suitable for deployment in real-world environments.


\section{Proofs of Theoretical Results}
\label{proofs}
To facilitate clearer understanding, we first provide a comprehensive overview of all notations used throughout the subsequent proofs, as summarized in \Cref{tab:symbol}. This notation table serves as a quick reference for the key variables, mappings, and geometric quantities involved in our analysis. Following this, we present detailed proofs for all Proposition introduced in the main paper, including the lower bound on raw reconstruction error and the lower bound under edit perturbation. These proofs offer additional mathematical clarity and support the theoretical foundations of EIRES.

\begin{table}[!t]
\centering
\begin{tabular}{@{}ll@{}}
\toprule
\textbf{Symbol} & \textbf{Meaning} \\ \midrule
$\mathcal{M}$ & Reconsrtucion manifold $\displaystyle\mathcal{M}=\{D(z)\mid z\in\mathbb{R}^{d}\}$ \\
$\mathcal{U}$ & Tubular neighbourhood of $\mathcal{M}$\\
$x_{r}$ & Query image, off-manifold $\mathcal M$ \\
$\tilde{x}_{r}$ & Nearest point of $x_{r}$ on $\mathcal{M}$, i.e., $\tilde{x}_{r}=\mathcal P(x_{r})$ \\
$T_{\tilde{x}_{r}}\mathcal{M}$ & Tangent space of $\mathcal{M}$ at $\tilde{x}_{r}$ \\
$\varepsilon_{\perp}$ & Normal deviation $\varepsilon_{\perp}=x_{r}-\tilde{x}_{r}$, $\varepsilon_{\perp}\!\perp\!T_{\tilde{x}_{r}}\mathcal{M}$ \\
$z^{\ast}$ & Latent code of $\tilde{x}_{r}$: $z^{\ast}=E(\tilde{x}_{r})$ \\
$J_{E}(x)$ & Jacobian of $E$ at $x$, a $d\times HW3$ matrix \\
$J_{D}(z)$ & Jacobian of $D$ at $z$, a $HW3\times d$ matrix \\
$\sigma_{\min}(\cdot)$ & Minimum singular value\\
$\sigma_{\max}(\cdot)$ & Maximum singular value \\
$\kappa_{D}$ & condition number $ \kappa_{D}={\sigma_{\max}\!\bigl(J_{D}\bigr)}/{\sigma_{\min}\!\bigl(J_{D}\bigr)}$ \\ 
\bottomrule
\end{tabular}
\caption{Symbols and notation used throughout the paper}
\label{tab:symbol}
\end{table}

\subsection{Lower Bound for Reconstruction Error}
 We  investigate the behavior of reconstruction error in the vicinity of the decoder’s reconstruction manifold. Our goal is to establish a theoretical lower bound on the reconstruction error by analyzing the decoder’s local geometric properties, specifically through the lens of its Jacobian condition number.
In particular, we model the decoder as implicitly defining a manifold $\mathcal{M}$ in the data space. For inputs near $\mathcal{M}$, projecting onto the manifold introduces a residual primarily along the normal direction. Since the decoder exhibits limited sensitivity to off-manifold perturbations, this normal residual is expected to dominate the reconstruction error. 
To formalize this intuition, we first present a lemma that characterizes the local structure of the decoder around $\mathcal{M}$, thereby laying the foundation for a principled lower bound on reconstruction error.

\begin{lemma}
\label{lem:41}
Let $E$ and $D$ be $\mathcal{C}^{1}$ maps satisfying $E \!\circ\! D = \operatorname{id}$ on the data manifold
$\mathcal{M}$. Then there exists an open neighbourhood $\mathcal{U}$ such that for any $x_{r}\in\mathcal{U}$, the nearest point on the manifold $\mathcal{M}$ is uniquely defined as 
\begin{equation}
\tilde{x}_{r} =\mathcal{P}(x_{r}) =
        \operatorname*{arg\,min}_{x\in\mathcal{M}}\,
        \|x_{r}-x\|_{2},
\end{equation}
and \(x_{r}\) can be decomposed as
\begin{align}
x_{r} = \tilde{x}_{r} + \varepsilon_{\perp},\;\;\varepsilon_{\perp}\;\perp\;T_{\tilde{x}_{r}}\mathcal{M},
\end{align}
where $T_{\tilde{x}_{r}}\mathcal{M}=\operatorname{Im}J_{D}\!\bigl(z^{\ast}\bigr)$, $z^{\ast}:=E(\tilde{x}_{r})$ and $J_{D}(z^{\ast})$ is the Jacobian of $D$ evaluated at $z^{\ast}$.
\end{lemma}

\begin{proof}
Since \( D \) is a \( \mathcal{C}^{1} \) map and satisfies the condition \( E \circ D = \operatorname{id} \) on the latent manifold \( \mathcal{M} \), the decoder \( D \) defines a regular, smooth embedding of a \( d \)-dimensional submanifold into the ambient space \( \mathbb{R}^{H \times W \times 3} \). According to the classical tubular neighborhood theorem for embedded submanifolds (\cite{lee2003smooth}, Theorem~10.19), there exists an open neighborhood \( \mathcal{U} \subset \mathbb{R}^{H \times W \times 3} \) containing \( \mathcal{M} \), together with a smooth map \( \mathcal{P} \colon \mathcal{U} \rightarrow \mathcal{M} \), referred to as the normal projection, such that for any point \( x_r \in \mathcal{U} \), the image \( \mathcal{P}(x_r) = \tilde{x}_r \in \mathcal{M} \) is the unique closest point to \( x_r \) on the manifold \( \mathcal{M} \) in terms of Euclidean distance. The difference vector \( \varepsilon_{\perp} = x_r - \mathcal{P}(x_r) \) lies in the normal bundle of \( \mathcal{M} \) at the point \( \tilde{x}_r \), and is therefore orthogonal to the tangent space \( T_{\tilde{x}_r} \mathcal{M} \). This orthogonality condition arises from the first-order optimality condition associated with Euclidean projection onto a smooth submanifold. Moreover, the smoothness of the projection map \( \mathcal{P} \) is guaranteed by the structure provided by the tubular neighborhood theorem.
\end{proof}

To streamline the proofs of Proposition~\ref{prop:41} and Proposition~\ref{prop:42}, we first introduce an auxiliary lemma. This lemma provides a key geometric relation that simplifies the subsequent derivations and allows for a more structured and tractable proof.

\begin{lemma}
\label{lem:42}
Let \( E \) and \( D \) be \( \mathcal{C}^1 \) maps that satisfy \( E \circ D = \operatorname{id} \) on the data manifold \( \mathcal{M} \). Then, for any \( x = D(z^{\ast}) \) with \( z^{\ast} = E(x) \), it holds that
\begin{equation}
\left\| J_E(x) \right\|_2 = \sigma_{\min}^{-1} \bigl( J_D(z^{\ast}) \bigr),
\end{equation}
where \( \sigma_{\min}(\cdot) \) denotes the smallest singular value.
\end{lemma}
\begin{proof}
Since \( E \circ D = \operatorname{id} \) on the data manifold \( \mathcal{M} \), we differentiate this identity at a point \( x = D(z^{\ast}) \), where \( z^{\ast} = E(x) \). By the chain rule, we obtain
\begin{align}
J_E(x)\, J_D(z^{\ast}) = I_d,
\end{align}
where \( I_d \) is the \( d \times d \) identity matrix. This shows that \( J_E(x) \) is a left inverse of \( J_D(z^{\ast}) \).

Now recall that for any full column-rank matrix \( A \in \mathbb{R}^{n \times d} \) with \( n \geq d \), the Moore--Penrose pseudoinverse \( A^{+} \) satisfies
\begin{align}
\| A^{+} \|_2 = \sigma_{\min}^{-1}(A),
\end{align}
and furthermore achieves the minimum operator 2-norm among all left inverses of \( A \)(~\cite{golub2013matrix}, Theorem~5.2). Applying this result to \( A = J_D(z^{\ast}) \), we conclude that any left inverse \( J_E(x) \) satisfies
\begin{align}
\| J_E(x) \|_2 \geq \| J_D(z^{\ast})^{+} \|_2 = \sigma_{\min}^{-1}( J_D(z^{\ast}) ).
\end{align}
But since \( J_E(x)\, J_D(z^{\ast}) = I_d \), the bound is attained and equality holds:
\begin{align}
\| J_E(x) \|_2 = \sigma_{\min}^{-1}( J_D(z^{\ast}) ).
\end{align}
\end{proof}



Based on the auxiliary lemma introduced above, we now provide the complete version of Proposition 1 from the main paper. The main text presents only a concise form of this result, and for completeness, the full statement is provided here.

\begin{proposition}[Detailed form of proposition~\textcolor{red}{1} in the main paper]
\label{prop:41}
Let $E$ and $D$ be $\mathcal{C}^{1}$ maps satisfying $E \circ D = \operatorname{id}$ on the data manifold $\mathcal{M}$. For any input $x \in \mathbb{R}^{H \times W \times 3}$, let $\tilde{x} \in \mathcal{M}$ denote the nearest point on the manifold, and define the normal deviation as $\varepsilon_\perp = x - \tilde{x}$. The reconstruction error behaves as follows:
\begin{itemize}
  \item If $x \in \mathcal{M}$, then $\varepsilon_\perp=0$,
  \begin{equation}
  \|x - D(E(x))\|_2 = \mathcal{O}(\|\varepsilon_\perp\|_2^2) \to 0.
  \label{eq:rein}
  \end{equation}
  \item If $x \notin \mathcal{M}$, then
  \begin{equation}
  \label{eq:re}
  \|x - D(E(x))\|_2 \geq \sqrt{1 + \kappa_D^{-2}} \|\varepsilon_\perp\|_2 + \mathcal{O}(\|\varepsilon_\perp\|_2^2),
  \end{equation}
  where $\kappa_D = \sigma_{\max}(J_D)/\sigma_{\min}(J_D)$ is the condition number of the decoder Jacobian evaluated at $\tilde{x}$.
\end{itemize}
\end{proposition}

\begin{proof}
Let $z^{\ast}=E(\tilde{x})$ and set $\Delta z := E(x)-z^{\ast}.$
A first-order Taylor expansion of \(E\) at \(\tilde{x}\) gives
\begin{align}
E\bigl(\tilde{x}+ \varepsilon_\perp\bigr)
  \;=\;
  E(\tilde{x})
  + J_E(\tilde{x})\,\varepsilon_\perp
  + \mathcal{O}\!\bigl(\,\|\varepsilon_\perp\|_2^{\,2}\bigr),
\end{align}
so that
\begin{align}
    \|\Delta z\|_2 \ge & \,\|J_E(\tilde{x})\,\varepsilon_\perp\|_2  - \|\mathcal{O}(\|\varepsilon_\perp\|_2^2)\|_2 \nonumber \\
    \ge\, & \sigma_{\min}(J_E(\tilde{x}))\,\|\varepsilon_\perp\|_2  + \mathcal{O}(\|\varepsilon_\perp\|_2^2).
\label{eq:A}
\end{align}
A first-order expansion of \(D\) at \(z^{\ast}\) yields
\begin{align}
    D(z^{\ast}+\Delta z)
  =  D(z^{\ast}) + J_D(z^{\ast})\,\Delta z
    + \mathcal O(\|\Delta z\|_2^{\,2}).
\end{align}
Hence
\begin{align}
x- D(E(x))
  = \varepsilon_\perp - J_D(z^{\ast})\,\Delta z
    + \mathcal O(\|\Delta z\|_2^{\,2}).
\label{eq:B}
\end{align}
Because \(J_D(z^{\ast})\,\Delta z \in T_{\tilde{x}}\mathcal M\) while
\(\varepsilon_\perp \perp T_{\tilde{x}}\mathcal M\),
the first two terms in \Cref{eq:B} are orthogonal, implying:
\begin{align}
  \|x-  D(E(x))\|_2^{2}&
  = \|\varepsilon_\perp\|_2^{2} \nonumber
   \\
  & + \|J_D(z^{\ast})\,\Delta z\|_2^{2}
    + \mathcal O(\|\Delta z\|_2^{\,3}).  
\end{align}
Lemma~\ref{lem:42} together with \Cref{eq:A} gives
\begin{align}
  \|J_D(z^{\ast})\,\Delta z\|_2
    \ge & \,\sigma_{\min}(J_D(z^{\ast}))\|\Delta z\|_2  \nonumber\\
    \ge &\, \kappa_{D} \|\varepsilon_\perp\|_2^{2} + \mathcal O(\|\varepsilon_\perp\|_2^{\,2}).
\end{align}
Taking square roots and absorbing high-order terms yields 
\begin{align}
    \|x- D(E(x))\|_2\ge
\sqrt{1 + \kappa_D^{-2}}\|\varepsilon_\perp\|_2
+ \mathcal{O}(\|\varepsilon_\perp\|_2^2),
\end{align}
then establishing \Cref{eq:re}.

If \(x= x_g\), i.e.\ \(\varepsilon_\perp = 0\) and $\Delta z =0$,
immediately gives \Cref{eq:rein}.
\end{proof}

\Cref{prop:41} provides strong theoretical support for using reconstruction error to distinguish real from generated images. Specifically, real images lie off the reconstruction manifold and exhibit a non-zero reconstruction error lower bounded by $\sqrt{1 + \kappa_D^{-2}} \|\varepsilon_\perp\|_2$, while generated images lie on the manifold with near-zero error. As a direct consequence of this theorem, any perturbation along the normal direction leads to a non-negligible change in reconstruction error, forming the basis for our detection method.

\subsection{Lower Bound for EIRES score}
\begin{proposition}[[Proposition 2 in the main paper] ]
\label{prop:42}
Under the same conditions as Proposition~\ref{prop:41}, let $x_{\mathcal{T}} = x + \delta$ denote a structured edit with $\|\delta\|$ sufficiently small.  
Let $\tilde{x}_{\mathcal{T}}$ be its manifold projection and $\varepsilon_{\perp}^{(\mathcal{T})}=x_{\mathcal{T}} - \tilde{x}_{\mathcal{T}}$ the new normal deviation.  
Then:
\begin{align}
    \|x_{\mathcal{T}} - D(E(x_{\mathcal{T}}))\|_2
\ge
\sqrt{1+\kappa_D^{-2}}\|\varepsilon_{\perp}^{\mathcal{T}}\|_2+o(\|\varepsilon_{\perp}^{\mathcal{T}}\|_2^2).
\end{align}
Moreover, the reconstruction error shift satisfies
\begin{align}
    \Delta(x)
    =e(x_{\mathcal{T}})-e(x)
    \propto
    \|\varepsilon_{\perp}^{(\mathcal{T})}\|_2 - \|\varepsilon_{\perp}\|_2.
    \vspace{-0.5em}
\end{align}
\end{proposition}

\begin{proof}
We begin by recalling the necessary conditions from Proposition~\ref{prop:41}. We assume that $x_{\mathcal{T}} = x + \delta$ denotes a structured edit, where $\|\delta\|$ is sufficiently small. Let $\tilde{x}_{\mathcal{T}}$ be its manifold projection, and define $\varepsilon_{\perp}^{(\mathcal{T})} = x_{\mathcal{T}} - \tilde{x}_{\mathcal{T}}$ as the new normal deviation.
From Lemma~\ref{lem:41} and Proposition ~\ref{prop:41}, we know that the reconstruction error satisfies:

\begin{align}
\| x_{\mathcal{T}} - D(E(x_{\mathcal{T}})) \|_2 \ge \sqrt{1 + \kappa_D^{-2}} \|\varepsilon_{\perp}^{(\mathcal{T})}\|_2 + o(\|\varepsilon_{\perp}^{(\mathcal{T})}\|_2^2).
\end{align}

This follows from the geometric analysis and the regularity of the manifold where $D(E(x_{\mathcal{T}}))$ represents the reconstruction of $x_{\mathcal{T}}$. The term $\kappa_D$ represents the curvature of the manifold at the point $x_{\mathcal{T}}$ and is derived from the Jacobian of the decoder, which gives the lower bound for the reconstruction error.

Now, we examine the shift in reconstruction error due to the structured edit. The shift $\Delta(x)$ in reconstruction error is given by:
\begin{align}
\Delta(x) = e(x_{\mathcal{T}}) - e(x),
\end{align}
where $e(x)$ represents the reconstruction error of the image $x$. According to the result from Lemma~\ref{lem:42} and Proposition~\ref{prop:42}, the shift in the reconstruction error due to the edit is proportional to the difference in the normal deviations between $x_{\mathcal{T}}$ and $x$:
\begin{align}
\Delta(x) \propto \|\varepsilon_{\perp}^{(\mathcal{T})}\|_2 - \|\varepsilon_{\perp}\|_2.
\end{align}

This result directly follows from the fact that the reconstruction error shift depends on how much the image deviates from the manifold after the edit. The term $\varepsilon_{\perp}$ represents the normal deviation of the original image $x$, while $\varepsilon_{\perp}^{(\mathcal{T})}$ represents the normal deviation of the edited image $x_{\mathcal{T}}$.

Combining these results, we conclude that:
\begin{align}
\| x_{\mathcal{T}} - D(E(x_{\mathcal{T}})) \|_2 \ge \sqrt{1 + \kappa_D^{-2}} \|\varepsilon_{\perp}^{(\mathcal{T})}\|_2 + o(\|\varepsilon_{\perp}^{(\mathcal{T})}\|_2^2),
\end{align}
and the reconstruction error shift satisfies:
\begin{align}
\Delta(x) \propto \|\varepsilon_{\perp}^{(\mathcal{T})}\|_2 - \|\varepsilon_{\perp}\|_2.
\end{align}
\end{proof}

Proposition~\ref{prop:42} provides a critical insight into how the reconstruction error behaves under structured edits. Specifically, it shows that the deviation of the edited image from the manifold, denoted as $\varepsilon_{\perp}^{(\mathcal{T})}$, plays a central role in determining the magnitude of the reconstruction error shift. This proposition emphasizes that the reconstruction error of the edited image $x_{\mathcal{T}}$ can be bounded in terms of the normal deviation from the manifold, which helps us understand the behavior of the reconstruction error in a geometrically rigorous manner. The relationship between $\|\varepsilon_{\perp}^{(\mathcal{T})}\|_2$ and the reconstruction error shift is crucial for establishing the lower bound and understanding the sensitivity of the model to structured edits.

\section{Limitations and Future Work}
\label{future}
While EIRES demonstrates strong separability and robustness across diverse generative models, several limitations remain. The method relies on the assumption that real and generated images respond differently to structured editing, yet this response gap may diminish in challenging scenarios such as heavily stylized content, severe domain shifts, or real images that are inherently difficult to reconstruct. Improving the stability of this response under such conditions may require adaptive editing strategies that automatically adjust perturbation strength, edit type, or spatial region based on the input image. Moreover, EIRES currently focuses on image-level detection. Extending the edit-induced reconstruction paradigm to tasks such as localizing manipulated regions, analyzing multimodal consistency, or detecting generated videos represents a promising direction. Exploring these broader applications may further enhance the generality and forensic value of edit-induced probing signals.